\documentclass[10pt,twocolumn,letterpaper]{article}

\usepackage{iccv}
\usepackage{amsmath}
\usepackage{amssymb}

\usepackage{times}
\usepackage{bbm}
\usepackage{bm}

\usepackage{graphicx} 
\usepackage{subfigure} 
\usepackage{color}
\definecolor{bblue}{RGB}{0,30,95}
\definecolor{rred}{RGB}{190,0,0}
\definecolor{mygray}{gray}{.9}
\definecolor{ggray}{RGB}{127,127,127}

\usepackage{colortbl} 
\usepackage{multirow}
\makeatletter
\newcommand{\thickhline}{%
    \noalign {\ifnum 0=`}\fi \hrule height 1pt
    \futurelet \reserved@a \@xhline
}
\makeatother

\usepackage{caption}
\usepackage[pagebackref=true,breaklinks=true,colorlinks=true,bookmarks=false]{hyperref} 
\usepackage[export]{adjustbox}
\usepackage{enumitem}

\usepackage{algorithm}
\usepackage{xcolor}
\usepackage{listings}

\usepackage[numbers,sort&compress]{natbib}

\newcommand*\printvalue[1]{\texttt{\string #1} : \the #1}
\newcommand{\makesupptitle}[1]{
	\twocolumn[
	\begin{center}
		{\Large \bf #1 \par}
		{
		\large
		\lineskip .5em
		\par
		}
		\vskip .5em
		\vspace*{12pt}
	\end{center}
	]
}

\usepackage[pagebackref=true,breaklinks=true,bookmarks=false]{hyperref}

\iccvfinalcopy 

\def\iccvPaperID{***} 

\ificcvfinal\fi

\begin{document}

\title{Bird’s-Eye-View Scene Graph for Vision-Language Navigation}
\author{{Rui Liu
\quad Xiaohan Wang
\quad Wenguan Wang\thanks{Corresponding author: Wenguan Wang.}
\quad Yi Yang} \\
\small{ReLER, CCAI, Zhejiang University} \\
\small \url{https://github.com/DefaultRui/BEV-Scene-Graph}
}
\maketitle
\ificcvfinal\thispagestyle{empty}\fi

\begin{abstract}
Vision-language navigation (VLN), which entails an agent to navigate 3D environments following human instructions, has shown great advances. However, current agents are built upon panoramic observations, which hinders their ability to perceive 3D scene geometry and easily leads to ambiguous selection of panoramic view. To address these limitations, we present a BEV Scene Graph (BSG), which leverages multi-step BEV representations to encode scene layouts and geometric cues of indoor environment under the supervision of 3D detection. During navigation, BSG builds a local BEV representation at each step and maintains a BEV-based global scene map, which stores and organizes all the online collected local BEV representations according to their topological relations. Based on BSG, the agent predicts a local BEV grid-level decision score and a global graph-level decision score, combined with a subview selection score on panoramic views, for more accurate action prediction. Our approach significantly outperforms state-of-the-art methods on REVERIE, R2R, and R4R, showing the potential of BEV perception in VLN.
\end{abstract}
\vspace{-5pt}
\section{Introduction}
\vspace{-3pt}
Vision-language navigation (VLN) task~\cite{AndersonWTB0S0G18} requires an agent to navigate through a 3D environment~\cite{chang2017matterport3d} to a target location, according to natural language instructions. Existing work has made great advances in cross-modal reasoning~\cite{FriedHCRAMBSKD18,wang2019reinforced,TanYB19,hong2020language,ZhuQNSBWWEW22,lin2022adapt}, path planning~\cite{MaLWAKSX19,anderson2019chasing,ke2019tactical,deng2020evolving,chen2021topological}, and auxiliary tasks for pretraining~\cite{MajumdarSLAPB20,HaoLLCG20,zhu2020vision,ChenGSL21,qiao2022hop}. Their core ideas are learning to relate the language instructions to panoramic images of the environment. Though straightforward, these approaches heavily rely on 2D panoramic observations. As a result, they lack the capacity to preserve scene layouts and 3D structure, which are critical for navigation decision-making in embodied scenes. Moreover, indoor environments~\cite{song2015sun,dai2017scannet,chang2017matterport3d,baruch1arkitscenes} are characterized by substantial occlusion~\cite{yang2019embodied,wijmans2019embodied,patil2023advances}, posing challenges for the agent to accurately identify the objects and landmarks referenced by the instructions~\cite{AndersonWTB0S0G18,qi2020reverie}.

\begin{figure}[t]
	\begin{center}
		\includegraphics[width=0.98\linewidth]{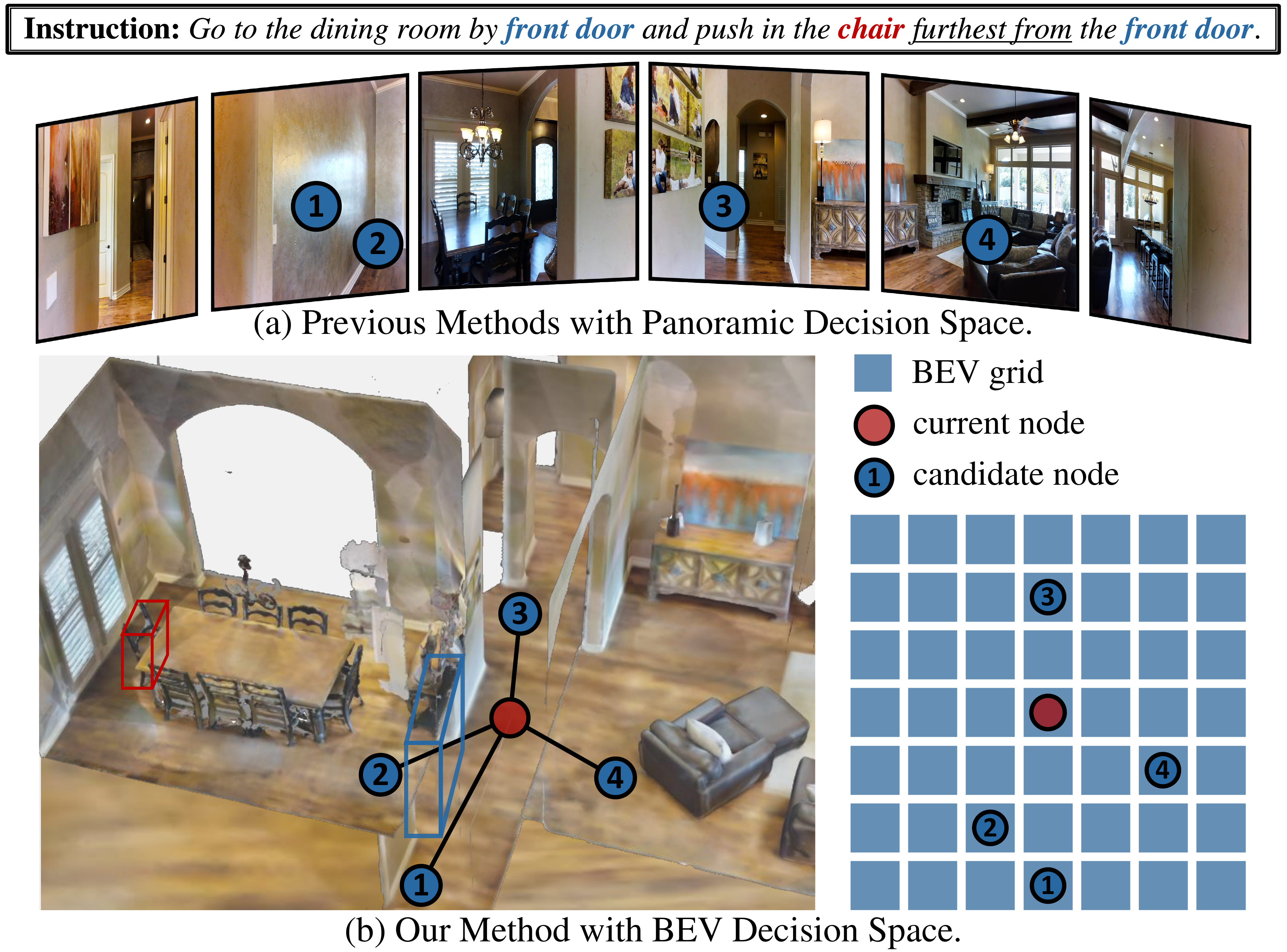}
	\end{center}
	\vspace{-16pt}
	\captionsetup{font=small}
	\caption{\small{For panoramic view (a),$_{\!}$ two candidate$_{\!}$ nodes$_{\!}$ (\protect\includegraphics[scale=0.15,valign=c]{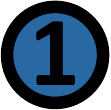}\&\protect\includegraphics[scale=0.15,valign=c]{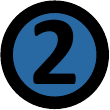})$_{\!}$ correspond to the same image$_{\!}$ leading to ambiguity. For Bird’s-Eye-View (b),$_{\!}$ they are represented$_{\!}$ by$_{\!}$ discriminative$_{\!}$ grids$_{\!}$ (\protect\includegraphics[scale=0.12,valign=c]{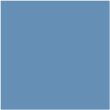}).}}
	\label{fig:introduce}
	\vspace{-15pt}
\end{figure}

For example (Fig.\!~\ref{fig:introduce}$_{\!}$(a)),$_{\!}$ given the$_{\!}$ instruction$_{\!}$ ``\textit{Go to the dining room by front door and push in the chair furthest from the front door}'',$_{\!}$ previous$_{\!}$ approaches$_{\!}$~\cite{FriedHCRAMBSKD18,wang2019reinforced,TanYB19,wang2020active,HaoLLCG20,MajumdarSLAPB20,ChenGSL21,moudgil2021soat,ChenGTSL22,GuhurTCLS21,Hong0QOG21} formulate VLN as a sequential text-to-image grounding problem by matching navigable candidate nodes with adjacent panoramic views. At each time step, given a set of subviews captured from different directions, the agent selects a navigable direction as the next step for navigation. However, this strategy tends to introduce ambiguity, when the agent needs to discriminate between multiple candidate nodes corresponding to the same subview. In addition, the agent struggles to ground the associated objects and explore their spatial relation in 3D scene, such as identifying ``\textit{the chair furthest from the front door}''. Consequently, relying solely on panoramic view presents difficulties in both comprehensive scene perception and efficient navigation.

\vspace{-1pt}
To address$_{\!}$ the challenges encountered$_{\!}$ by panoramic methods,$_{\!}$ Bird’s-Eye-View (BEV) perception emerges as a viable solution, employing discriminative grid representations to model the 3D environment. Meanwhile, BEV grid representation effectively captures spatial context$_{\!}$ and$_{\!}$ scene$_{\!}$ layouts~\cite{li2022delving,ma2022vision},$_{\!}$ facilitating both$_{\!}$ perception$_{\!}$ \cite{huang2021bevdet,li2022bevdepth,li2022bevformer,huang2023geometric} and planning~\cite{zhang2021end,hu2022stp3,zhao2022jperceiver,hu2022goal}.$_{\!}$ Building upon these insights, we present$_{\!}$ a BEV$_{\!}$ Scene$_{\!}$ Graph (BSG), which harnesses the power of BEV representation to construct an informative navigation graph. During navigation, the agent collects local BEV representations at each navigable node. A$_{\!}$ global$_{\!}$ scene$_{\!}$ graph$_{\!}$ is$_{\!}$ established by connecting these$_{\!}$ BEV$_{\!}$ representations$_{\!}$ topologically. At each step, the agent makes an informed decision by predicting a BEV grid-level decision score and a BSG graph-level decision score, combined with a subview selection score on panoramic views~\cite{ChenGTSL22,chen2021topological,wang2021structured}.

Specifically, the agent acquires multi-view observations at each$_{\!}$ step and performs view$_{\!}$ transformation~\cite{philion2020lift,reading2021categorical,wang2022detr3d,li2022bevformer} on the corresponding image features. Later, a 3D detection head$_{\!}$~\cite{stewart2016end,carion2020end,wang2022detr3d} is employed on these BEV representations to predict oriented bounding boxes, encoding object-level geometric and semantic information. During navigation, the node embeddings of BSG are represented by neighboring BEV grids. Then they are updated by querying the overlap region between BEV representations from different steps.

Previous semantic maps in robot navigation, including occupancy grids~\cite{elfes1990occupancy,ChenGG19,chaplot2020learning,ramakrishnan2021exploration} and learnable spatio-semantic representations~\cite{HenriquesV18,CartillierRJLEB21,georgakis2022cross,chen2022weakly,an2022bevbert}, have only provided top-down information without crucial 3D object information.$_{\!}$ Differently, BSG leverages the BEV representations to achieve consistency between 3D perception and decision-making while encoding geometric context. Our approach is evaluated on three benchmarks (\ie, REVERIE~\cite{qi2020reverie}, R2R~\cite{AndersonWTB0S0G18}, R4R~\cite{jain2019stay}). For the referring expression comprehension in REVERIE, BSG outperforms the state-of-the-art method~\cite{ChenGTSL22} by $\bm{5.14\%}$ and $\bm{3.21\%}$ in SR and RGS on the val unseen split, respectively. BSG also achieves $\bm{4\%}$ and $\bm{3\%}$ improvement in SR and SPL on the test split of R2R, respectively.$_{\!}$ The impressive results$_{\!}$ shed$_{\!}$ light on the promises$_{\!}$ of BEV perception$_{\!}$ in VLN$_{\!}$ task.

\section{Related Work}
\vspace{-3pt}
\noindent\textbf{Vision-Language Navigation (VLN).} VLN task~\cite{AndersonWTB0S0G18} has drawn significant attention in embodied AI domain. Early work typically adopts recurrent neural networks with cross-modal attention~\cite{AndersonWTB0S0G18,FriedHCRAMBSKD18,TanYB19,MaWAXK19,vasudevan2021talk2nav}. Later, various techniques have been developed to improve VLN, including: \textbf{i)} using more powerful vision-and-language embedding methods based on pre-trained transformer models~\cite{HaoLLCG20,LiLXBCGSC19,DevlinCLT19,MajumdarSLAPB20,GuhurTCLS21,Hong0QOG21,ChenGSL21,PashevichS021,ChenGTSL22,qiao2022hop}; \textbf{ii)} exploiting more supervisory information from environment augmentation~\cite{liu2021vision,li2022envedit,koh2021pathdreamer}, instruction generation~\cite{TanYB19,FriedHCRAMBSKD18,fu2020counterfactual,wang2022counterfactual,wang2023lana}, and other auxiliary tasks~\cite{wang2019reinforced,huang2019transferable,MaLWAKSX19,zhu2020vision,AnQHWWT21}; \textbf{iii)} designing more efficient action planning and learning strategies by incorporating self-correction~\cite{ke2019tactical,MaWAXK19}, global action space~\cite{wang2020active,wang2021structured,deng2020evolving,zhu2021soon,zhao2022target}, map building~\cite{ChenGTSL22,an2022bevbert,chen2021topological,wang2021structured}, knowledge prompts~\cite{lin2022adapt,yang2021multiple,li2023kerm},$_{\!}$ or$_{\!}$ ensemble$_{\!}$ of$_{\!}$ IL~\cite{AndersonWTB0S0G18}$_{\!}$ and$_{\!}$ RL$_{\!}$~\cite{wang2018look,wang2019reinforced};$_{\!}$ and$_{\!}$ \textbf{iv)} developing$_{\!}$ more$_{\!}$ large-scale benchmarks~\cite{qi2020reverie,zhu2021soon,ku2020room,krantz2020beyond,xia2018gibson,wang2022towards,chang2017matterport3d,savva2019habitat,RamakrishnanGWM21,szot2021habitat} and platforms$_{\!}$~\cite{chang2017matterport3d,savva2019habitat,RamakrishnanGWM21,szot2021habitat}.

However, existing work heavily relies on panoramic subviews for navigation, suffering from the limitations of 2D perspective view. These limitations, including occlusion and a narrow field of subview, introduce ambiguity in action prediction, thereby hindering efficient navigation. In contrast, we leverage BEV representations to facilitate navigation decision-making through view transformation. These BEV representations encode geometric context of environment under the supervision of BEV-based 3D detection.

\noindent\textbf{Map Representation for Navigation.} To achieve accurate navigation, it is critical to develop an efficient representation of surrounding environments. In robot navigation, classical SLAM-based approaches build a map based on geometry and plan the path on this semantic-agnostic map~\cite{thrun2002probabilistic,thrun1998learning,ChenGG19,ramakrishnan2021exploration}. These approaches are built upon sensors and thus highly susceptible to measurement noises~\cite{chaplot2020learning,chen2022weakly}. To explore semantic information, learnable semantic map~\cite{chen2022weakly,irshad2022semantically,anderson2019chasing,HenriquesV18,CartillierRJLEB21,elfes1990occupancy,ChenGG19,chaplot2020learning,ramakrishnan2021exploration} is proposed using the learnable spatial representations from a top-down view. These two types of metric maps focus on dense representations with explicit location information of environment.$_{\!}$ Moreover,$_{\!}$ topological maps~\cite{ChenGTSL22,chen2021topological,wang2021structured,deng2020evolving} are developed to model the relationship among sparse nodes in the environment, mitigating the burden of heavy computation. In addition, some efforts build topo-metric maps to combine the advantages of metric and topological maps~\cite{blanco2008toward,gomez2020hybrid,niijima2020city,an2022bevbert}.

Existing map-based methods neglect the role of 3D perception for$_{\!}$ navigation.$_{\!}$ In$_{\!}$ contrast,$_{\!}$ BSG encodes$_{\!}$ scene layouts and geometric$_{\!}$ cues$_{\!}$ by 3D detection$_{\!}$ for comprehensive scene$_{\!}$ understanding, eventually$_{\!}$ facilitating$_{\!}$ path$_{\!}$ planning.

\begin{figure*}[t]
	\begin{center}

		\includegraphics[width=0.98\linewidth]{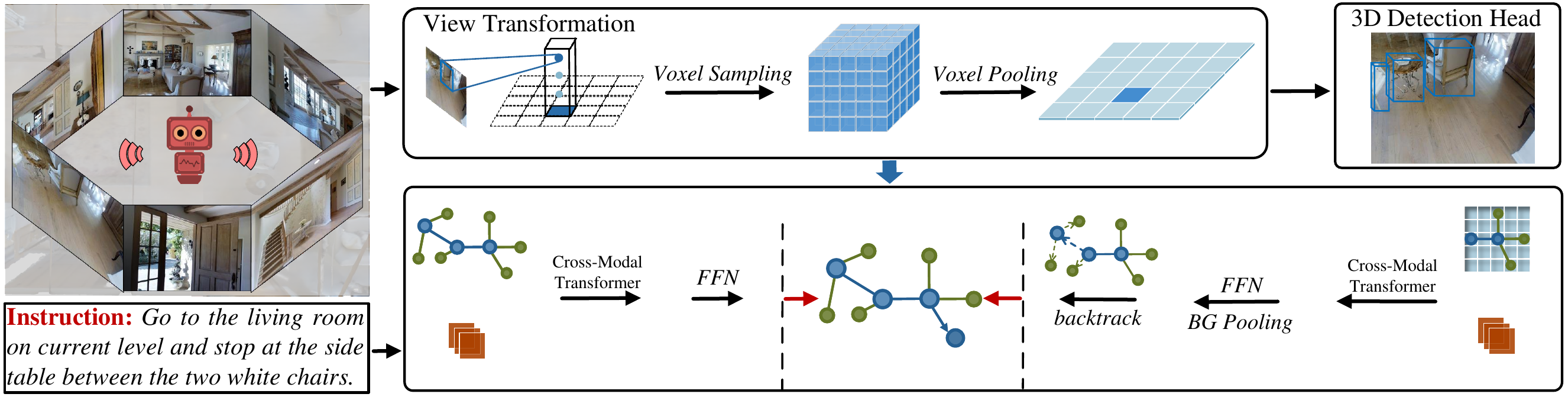}
        \put(-354,77){\footnotesize{$\bm{F}_{2D}$}}
        \put(-334,77){\footnotesize{$\bm{P}(h,w,z)$}}
        \put(-230,76){\footnotesize{$\bm{F}_{3D}$}}
        \put(-276,85){\footnotesize{Eq.\!~\ref{equ:voxelsam}}}
        \put(-166,77){\scriptsize{$\bm{B}_t\!=\!\{\bm{b}_i\}_{i=1}^{HW}$}}
        \put(-206,67){\scriptsize{$\bm{B}_t$}}
        \put(-360,28){\scriptsize{$\bm{V}^{\!g}\!\!{_{\!}}=\!\!{_{\!}}\{{_{\!}}\bm{V}_i{_{\!}}\}_{{_{\!}}{i=1}}^{{_{\!\!}}{N_t}{_{\!}}+{_{\!}}1}$}}
        \put(-346,4.5){\footnotesize{$\bm{X}$}}
        \put(-284,28){\footnotesize{$\tilde{\bm{V}}^{\!g}$}}
        \put(-344,58){\small{Graph-Level Decision Score}}
        \put(-311,23){\scriptsize{Eq.\!~\ref{equ:crossgraph}}}
        \put(-253,28){\footnotesize{${\bm s}^n$}}
        \put(-274,23){\scriptsize{Eq.\!~\ref{equ:globalscore}}}
        \put(-216,8){\scriptsize{Eq.\!~\ref{equ:globallocal}}}
        \put(-166,28){\footnotesize{${\hat{s}}^c$}}
        \put(-156,13){\scriptsize{Eq.\!~\ref{equ:backtrack}}}
        \put(-127,28){\footnotesize{$s^c$}}
        \put(-116,13){\scriptsize{Eq.\!~\ref{equ:gridscore}\&\ref{equ:grid2node}}}
        \put(-85,27){\footnotesize{$\tilde{\bm{B}}_t$}}
        \put(-64,22){\scriptsize{Eq.\!~\ref{equ:crossbev}}}
        \put(-27,5){\footnotesize{$\bm{X}$}}
        \put(-28,30){\footnotesize{${\bm{B}}_t$}}
        \put(-133,57.5){\small{Grid-Level Decision Score}}
        \put(-227,57.5){\small{Fused Score}}
	\end{center}
	\vspace{-13pt}
	\captionsetup{font=small}
	\caption{\small{Overview of BSG. View transformation is first employed to project the multi-view images into BEV plane (\S\ref{sec:BSG}). Then, BEV feature is encoded using 3D detection (\S\ref{sec:encodebev}). Through the integration of BEV representations during navigation, we predict a graph-level decision score on BSG and a grid-level decision score based on BEV. These scores are fused to facilitate effective decision-making (\S\ref{sec:decision}).}}
	\label{fig:framework}
	\vspace{-2pt}
\end{figure*}
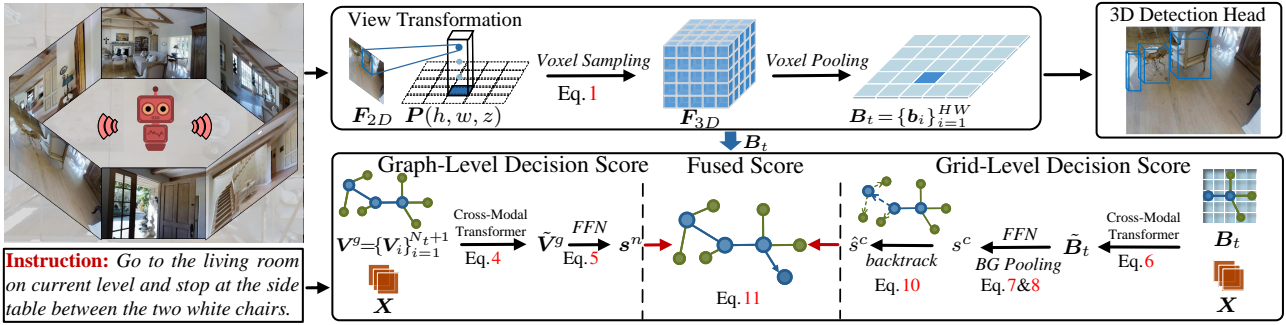

\noindent\textbf{Perceptual Organization of 3D Scenes.} Scene representation should provide information about both object semantics and layout composition~\cite{gupta2013perceptual,koppula2011semantic,mildenhall2020nerf,jiang2020local,WuLHZ021,niemeyer2020differentiable,patil2023advances}. For indoor scene understanding, visual representation can take various forms, including an RGB image and depth map~\cite{dai2017scannet,song2015sun,baruch1arkitscenes}, voxel grids~\cite{song2017semantic}, and point clouds~\cite{qi2017pointnet,qi2017pointnet++}. As pointed by~\cite{patil2023advances}, structural representation~\cite{chaudhuri2020learning,hong2020language,wald2020learning} also plays a significant role, as it models the spatial relationships among different objects. Therefore, modeling visual and structural properties is critical for scene understanding. Recently, BEV feature provides a unified representation for perception and motion planning~\cite{zhang2021end,hu2022stp3,zhao2022jperceiver,hu2022goal,li2022delving,ma2022vision}.

Motivated by the recent efforts that achieve learnable projection between BEV plane and perspective view~\cite{philion2020lift,reading2021categorical,huang2021bevdet,li2022bevdepth,liu2022bevfusion,liang2022bevfusion,park2021pseudo,guo2021liga,chen2020dsgn,li2022bevformer}, we collect oriented 3D bounding boxes in Matterport3D dataset~\cite{chang2017matterport3d} and perform camera-based BEV perception for embodied amodal detection~\cite{yang2019embodied,zhan2022tri}, as opposed to previous point cloud-based detection~\cite{pan20213d,armeni20163d,qi2018frustum}. Under the supervision of 3D detection, we employ BEV feature to establish scene representations that effectively capture object-level geometry information for navigation.

\section{Approach}\label{sec:approach}
\vspace{-3pt}
\noindent\textbf{Task Setup.} We illustrate our approach using R2R~\cite{AndersonWTB0S0G18} task, where the environment is discretized as a set of navigable nodes and navigability edges. The agent observes the surroundings at each node and finds a route to the target location, specified by the instruction $\mathcal{X}\!=\!\{x_l\}_{l=1}^{L}$ with $L$ words.

\noindent\textbf{Panoramic Methods.} Previous VLN agents~\cite{TanYB19,wang2019reinforced,Hong0QOG21,ChenGTSL22} are$_{\!}$ built$_{\!}$ as$_{\!}$ panoramic$_{\!}$ view$_{\!}$ selectors$_{\!}$~\cite{FriedHCRAMBSKD18}$_{\!}$ where$_{\!}$ navigable candidate nodes are represented by adjacent observations from different viewing angles. However, the adjacency rule in panoramic navigation will cause multiple candidate nodes to correspond to the same panoramic view, thus introducing ambiguity in action prediction (Fig.$_{\!}$ \ref{fig:introduce}(a)).$_{\!}$ In$_{\!}$ addition,$_{\!}$ the geometric$_{\!}$ cues$_{\!}$ of 3D environment$_{\!}$ cannot be$_{\!}$ captured$_{\!}$ by visual features of$_{\!}$ 2D panoramic views,$_{\!}$ such as occluded objects~\cite{wilkes1992active,palmer1999vision,yang2019embodied,zhan2022tri} and scene layouts~\cite{tsai2022towards,zhao2022jperceiver}.

\noindent\textbf{Our Idea.} To overcome the above limitations, we utilize BEV features as geometry-enhanced visual representations, supervised by BEV-based 3D detection. Then, we construct BEV Scene Graph (BSG) online using BEV features (Fig.~\ref{fig:introduce}(b)). With BSG, the agent effectively predicts the next step on candidate nodes, which are represented by discriminate BEV grids. Before detailing BEV detection (\S\ref{sec:encodebev}), we first introduce how to build BSG (\S\ref{sec:BSG}) and how to predict decision score for action prediction(\S\ref{sec:decision}).

\subsection{BSG Construction}\label{sec:BSG}
\begin{figure*}[t]
	\begin{center}
		\includegraphics[width=0.999\linewidth]{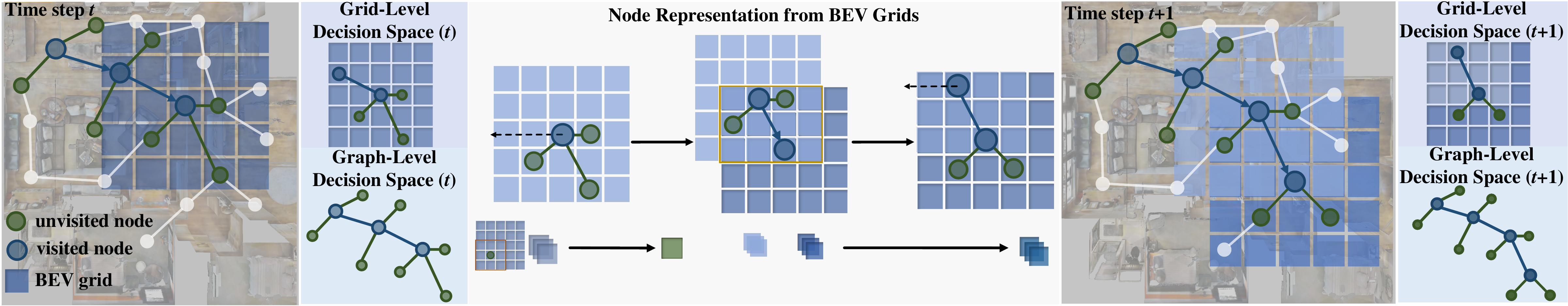}
        \put(-74,6){\scriptsize{${\bm{B}}_{t\!+\!1}$}}
        \put(-70,79){\scriptsize{${\bm{B}}_t$}}
        \put(-411,79){\scriptsize{${\bm{B}}_t$}}
        \put(-349,51){\scriptsize{${\bm{V}}_t$}}
        \put(-321,77){\scriptsize{${\bm{B}}_t$}}
        \put(-221,66){\scriptsize{${\bm{V}}_{t\!+\!1}$}}
        \put(-187,76){\scriptsize{${\bm{B}}_{t\!+\!1}$}}
        \put(-277,38){\scriptsize{$\Omega^{\rm{o}}$}}
        \put(-294,44){\scriptsize{Eq. \ref{equ:avegrid}}}
        \put(-224,44){\scriptsize{Eq. \ref{equ:bevupdate}}}
        \put(-313,21){\scriptsize{Average}}
        \put(-309,12){\scriptsize{Eq. \ref{equ:avegrid}}}
        \put(-342,3){\scriptsize{$\Omega^{\rm{n}}$}}
        \put(-328,3){\scriptsize{$\{{\bm b}_i\}_{i\in\Omega^{\rm{n}}}$}}
        \put(-287,3){\scriptsize{${\bm{V}}$}}
        \put(-267,3){\scriptsize{$\{{{\bm{b}}_{i,t}}\}$}}
        \put(-246,3){\scriptsize{$\{{{\bm{b}}_{j,t\!+\!1}}\}_{{i,j}\in\Omega^{\rm{o}}}$}}
        \put(-185,3){\scriptsize{$\{\!\tilde{\bm{b}}_{j,t\!+\!1}$\!\}}}
        \put(-210,11){\scriptsize{Eq. \ref{equ:bevupdate}}}
        \put(-225,20){\scriptsize{Cross-Attention}}
	\end{center}
	\vspace{-10pt}
	\captionsetup{font=small}
	\caption{\small{The node embeddings of BSG are represented by BEV grids in their neighborhood. From step \textit{t} to \textit{t}+1, BSG is updated using temporal modeling (\S\ref{sec:BSG}). Both global graph-level and local grid-level decision space are also used for accurate action prediction (\S\ref{sec:decision}).}}
	\label{fig:temporalagg}
	\vspace{-10pt}
\end{figure*}
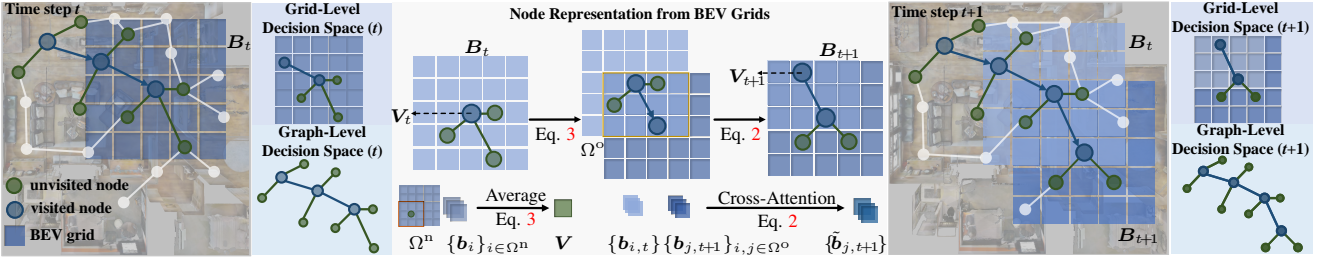

During navigation, the agent collects local BEV representations of surrounding environment online, and constructs a global scene graph gradually. Specifically, at time step $t$, BSG is denoted as $\mathcal{G}_t=\{\mathcal{V}_t,\mathcal{E}_t\}$, where each node $v\in\mathcal{V}_t$ incorporates observed information (Fig.\!~\ref{fig:temporalagg}), corresponding to each navigable location in the environment.

\noindent\textbf{View Transformation.}$_{\!}$ At$_{\!}$ current location$_{\!}$ $v^*$,$_{\!}$ the$_{\!}$ agent$_{\!}$ acquires$_{\!}$ multi-view$_{\!}$ camera$_{\!}$ images\footnote{As there are no specific camera parameters available for panoramic images from the simulator~\cite{AndersonWTB0S0G18}, we utilize the images captured by raw camera with intrinsic and extrinsic parameters~\cite{chang2017matterport3d}. Both types of images encompass identical visual content (see \S\ref{sec:annotation} for details).}.$_{\!}$ We$_{\!}$ perform \textit{voxel sampling}~\cite{wang2022detr3d,liu2022petr,li2022bevformer,li2022delving,ma2022vision}$_{\!}$ on$_{\!}$ each$_{\!}$ image$_{\!}$ feature$_{\!}$ $\bm{F}_{2D}\!\in\!\mathbb{R}^{{H_c}_{\!}{W_c}\!\times\!D}$$_{\!}$ to$_{\!}$ construct$_{\!}$ 3D$_{\!}$ voxel$_{\!}$ feature$_{\!}$ $\bm{F}_{3D}\!\!\in\!\!\mathbb{R}^{{H}\!{W}\!{Z}\!\times\!D}$, where ${H_c}{W_c}$ and $HW$ are the spatial dimensions of image feature and BEV plane, respectively. Predefined 3D reference points $\bm{P}\!\in\!\mathbb{R}^{{H}\!{W}\!{Z}}$ are used to query the image feature via \textit{cross-attention} for voxel feature (Fig.\!~\ref{fig:framework}),\! where ${H}{W}\!{Z}$ denotes the number of reference points:
        \vspace{-0pt}
        \begin{equation}
        \small
        \begin{aligned}
        \bm{F}_{3D}(h,w,z)={\text{CrossAtt}}\big(\bm{P}(h,w,z),{\bm{F}_{2D}(h_i,w_i)}\big).
        \end{aligned}
        \label{equ:voxelsam}
        \vspace{-0pt}
        \end{equation}

Then,$_{\!}$ $\bm{F}_{3D}$$_{\!}$ is$_{\!}$ squeezed$_{\!}$ down$_{\!}$ to BEV space by \textit{voxel pooling} as $\bm{B}\!\!=\!\!\{\bm{b}_i\}_{i=1}^{HW}\!\!\in\!\!\mathbb{R}^{{HW}\!\times\!D}$, where each grid cell contains$_{\!}$ a$_{\!}$ $D$-sized$_{\!}$ latent$_{\!}$ vector,$_{\!}$ representing$_{\!}$ the$_{\!}$ corresponding$_{\!}$ region$_{\!}$ in$_{\!}$ environment.$_{\!}$ Then, BEV feature is connected with a 3D detection head (\textit{cf}.~\!\S\ref{sec:encodebev}) to predict bounding boxes, providing the agent with object-level geometry information.

\noindent\textbf{Node Representation from BEV Grids.} At the start of navigation (\ie, $t\!=\!0$), BSG $\mathcal{G}_{0}$ is initialized with the node set $\mathcal{V}_0$ and its$_{\!}$ associated BEV feature$_{\!}$ $\bm{B}_0$ (Fig.\!~\ref{fig:temporalagg}). It is noted that there is an overlapping region $\Omega^{\rm{o}}$ between $\bm{B}_{t}$ and $\bm{B}_{t+1}$, since the perception range is greater than the moving step. At time step $t\!+\!1$, the same spatial region will be captured by different BEV grid features from $\Omega^{\rm{o}}$. Then, we execute temporal modeling on $\bm{B}_{t}$ and $\bm{B}_{t+1}$ to integrate history information, thereby facilitating the representation of stationary objects~\cite{li2022bevformer,huang2022bevdet4d,ma20223d}. In particular, we adopt \textit{cross-attention}~\cite{vaswani2017attention} on the grid features to update $\bm{B}_{t+1}$:
\vspace{1pt}
\begin{equation}
\small
\begin{aligned}
\tilde{\bm{b}}_{j,t+1}&={\text{CrossAtt}}\big({\bm{b}_{i,t}},{\bm{b}_{j,t+1}}\big),~i,j\in\Omega^{\rm{o}}.\\
\end{aligned}
\label{equ:bevupdate}
\vspace{1pt}
\end{equation}

Since$_{\!}$ local$_{\!}$ scene information is captured by corresponding BEV features, we construct node representations of BSG by incorporating the features of surrounding BEV grids, which are identified by nearest neighbor search~\cite{wang2019dynamic,wang2021object}. At step $t\!+\!1$, for current node $v^*$ and its navigable candidate nodes $\{v^+_k\}_{k=1}^{K_{t+1}}\!\!\!\in\!\!\mathcal{V}_{t+1}$, we$_{\!}$ \textit{average}$_{\!}$ the BEV grid$_{\!}$ features$_{\!}$ of corresponding neighborhood ${\Omega}^{\rm{n}}_{t}$:
\begin{equation}
\small
\begin{aligned}
\bm{V}_{t+1}={\text{Ave}}\big(\{\bm{b}_{i,t+1}\}_{i\in\Omega^{\rm{n}}_{t}}\big).
\end{aligned}
\label{equ:avegrid}
\vspace{2pt}
\end{equation}
Each node representation $\bm{V}_{t+1}\!\in\!\mathbb{R}^{D}$ attends to a certain area. For the candidate nodes that have been observed (or visited) multiple times, we \textit{average} the previous representations as its node embedding~\cite{wang2021structured,ChenGTSL22}. After updating BSG, we preserve  $\bm{B}_{t+1}$ for subsequent action prediction (\S\ref{sec:decision}).

\subsection{BEV-based Navigation Action Prediction}\label{sec:decision}
With the current BSG $\mathcal{G}_t\!=\!\{\mathcal{V}_t,\mathcal{E}_t\}$ and navigation instruction $\mathcal{X}$, the agent predicts next step by combining grid-level decision score on BEV feature $\bm{B}_t$ and graph-level decision score on BSG $\mathcal{G}_t$. Following~\cite{deng2020evolving,ChenGTSL22}, we add a hallucination ``stop'' node to existing $N_t$ ($=\!|\mathcal{V}_t|$) nodes.

\noindent\textbf{BSG-based Graph-level Decision Score.}$_{\!}$ The$_{\!}$ word$_{\!}$ embeddings $\bm{X}\!\!\in\!\!\mathbb{R}^{L\!\times\!D}$ and$_{\!}$ node$_{\!}$ embeddings $\bm{V}^g\!\!=\!\!\{\bm{V}_n\}_{n=1}^{{N_t}+1}\!\!\in\!\!\mathbb{R}^{(N_t+1)\!\times\!D}$ are fed into a \textit{cross-modal} encoder~\cite{tan2019lxmert} with several \textit{cross-attention} and \textit{self-attention} layers to model the relations between instruction and graph representations:
\vspace{1pt}
\begin{equation}
\small
\begin{aligned}
\tilde{\bm{V}}^g\!=\!\{\tilde{\bm{V}}_n\}_{n=1}^{{N_t}+1}\!=\!{\text{CrossMod}}\big([{\bm{V}^g,\bm{X}}]\big),
\end{aligned}
\label{equ:crossgraph}
\vspace{2pt}
\end{equation}
where $[\cdot]$ indicates the concatenation operation. After that, we adopt a feed-forward network (\textit{FFN}) to predict the global graph-level decision score $s^n\in\mathbb{R}^{{N_t}+1}$ of $\mathcal{G}_t$:
\vspace{-0pt}
\begin{equation}
\small
\begin{aligned}
{\bm s}^g\!=\!\{s^g_n\}_{n=1}^{{N_t}+1}\!=\!{\text{FFN}}\big(\tilde{\bm{V}}^g\big).
\end{aligned}
\label{equ:globalscore}
\vspace{-0pt}
\end{equation}

\noindent\textbf{BEV-based Grid-Level Decision Score.} Grid-level decision score on $\bm{B}_{t}$ is crucial for the agent to understand the 3D scene and learn effective navigation policies. A similar \textit{cross-modal} transformer~\cite{tan2019lxmert} is used to mine fine-grained visual clues and object-related textual information from the instructions, such as ``\textit{front door}'' and ``\textit{the chair furthest from the front door}'':
\vspace{-1pt}
\begin{equation}
\small
\begin{aligned}
\tilde{\bm{B}}_t\!=\!\{\tilde{\bm{b}}_{i}\}_{i=1}^{HW}\!=\!{\text{CrossMod}}\big([{\bm{B}_t,\bm{X}}]\big).
\end{aligned}
\label{equ:crossbev}
\vspace{-1pt}
\end{equation}

Then$_{\!}$ the$_{\!}$ instruction-aware representations $\tilde{\bm{B}}_t$ is used to predict local grid-level decision score $s^l\!\in\!\mathbb{R}^{HW}$ by \textit{FFN}:
\vspace{-1pt}
\begin{equation}
\small
\begin{aligned}
{\bm s}^l\!=\!\{s_i^l\}_{i=1}^{HW}\!=\!{\text{FFN}}\big(\tilde{\bm{B}}_t\big).
\end{aligned}
\label{equ:gridscore}
\vspace{-1pt}
\end{equation}

We propose a distance-dependent weighted pooling to convert the grid-level score $s^l$ to local candidate score $s^c\in\mathbb{R}^{K_t+1}$ (containing the stop node)~\cite{AndersonWTB0S0G18,qi2020reverie}. For $k$-th navigable candidate node, the score is calculated as follows:
\vspace{-1pt}
\begin{equation}
\small
\begin{aligned}
s^c_k=\sum\nolimits_{i\in\Omega^{\rm n}_k}{W_{k,i}}{s^l_i},
\end{aligned}
\vspace{-1pt}
\label{equ:grid2node}
\end{equation}
where $\Omega^{\rm n}_k$ is the grid neighborhood of $k$-th candidate node (\textit{cf}.$_{\!}$ Eq.$_{\!}$~\ref{equ:avegrid}),$_{\!}$ and$_{\!}$ ${\bm W}_{\!k}\!\!=\!\![W_{\!k,i}]_{i=1}^{|\Omega^{\rm n}_k|}$ is$_{\!}$ a$_{\!}$ truncated$_{\!}$ \textit{Bivariate Gaussian}$_{\!}$ weight,$_{\!}$ as$_{\!}$ the$_{\!}$ contribution$_{\!}$ of$_{\!}$ BEV$_{\!}$ grids$_{\!}$ to$_{\!}$ candidate$_{\!}$ nodes$_{\!}$ is$_{\!}$ considered$_{\!}$ contingent$_{\!}$ on$_{\!}$ relative$_{\!}$ distance:
\begin{equation}
\small
\begin{aligned}
W_{k,i}=\hat{g}(\Delta{x_{k,i}},\Delta{y_{k,i}}),
\end{aligned}
\vspace{-1pt}
\label{equ:gaussw}
\end{equation}
where $(\Delta{x_{k,i}},\Delta{y_{k,i}})$ is the relative coordinates of the $i$-th BEV grid center to $k$-th candidate node coordinates, $\hat{g}(\cdot)$ is normalized Bivariate Gaussian probability $\mathcal{N}(\bm{\mu}_{x,y},\bm{\sigma}_{x,y})$, $\bm{\mu}_{x,y}$ is the mean vector, and $\bm{\sigma}_{x,y}$ is the covariance matrix.

\noindent\textbf{Fused Action Prediction.} To fuse the global graph-level decision score and local grid-level decision score, a backtracking strategy~\cite{deng2020evolving,ChenGTSL22} is adopted to convert the local score $s^c\in\mathbb{R}^{K_t+1}$ into global space $\hat{s}^c\in\mathbb{R}^{N_t+1}$. Specifically, when navigating to the nodes that are not connected to the current node, we assume the agent needs to backtrack through the visited candidate nodes as:
\vspace{-1pt}
\begin{equation}
\small
\hat{s}^c=\left\{
\begin{aligned}
s_{\rm back},&~~~\rm {if~backtrack},\\
s^c,&~~~\rm otherwise.
\end{aligned}
\right.
\vspace{-1pt}
\label{equ:backtrack}
\end{equation}

More specifically, we compute a backtracking score for unconnected nodes in $\mathcal{V}_{t}$ by summing the decision scores of visited candidate nodes as$_{\!}$ $s_{\rm back}$. Then, a weight $W_f$ is employed to fuse the local and global decision scores:
\vspace{-1pt}
\begin{equation}
\small
\begin{aligned}
s_n={W_f}{\hat{s}}^c_n+(1-W_f)s^g_n.
\end{aligned}
\vspace{-1pt}
\label{equ:globallocal}
\end{equation}

Using the$_{\!}$ fused prediction,$_{\!}$ BSG$_{\!}$ can$_{\!}$ complement$_{\!}$ existing$_{\!}$ works$_{\!}$~\cite{deng2020evolving,wang2021structured,ChenGTSL22} with global$_{\!}$ action$_{\!}$ space.$_{\!}$ We will adopt a previous method~\cite{ChenGTSL22}$_{\!}$ as$_{\!}$ basic agent for experiment (\textit{cf}. \S\ref{sec:experiment}).

\subsection{BEV Representation Encoding}\label{sec:encodebev}
BEV detection endows the agent with awareness of object-level geometry information, facilitating more accurate action prediction~\cite{shen2019situational,tan2022self}. In this section, we learn 3D object detection on the top of BEV feature (see \S\ref{sec:BSG})~\cite{huang2021bevdet,li2022bevdepth,li2022bevformer}. Accordingly, the details on collecting a Matterport3D-based detection dataset for embodied amodal perception~\cite{yang2019embodied,wijmans2019embodied,nilsson2021embodied}, called ${\text{{Matterport3D}}}^2$, will be presented. We also introduce the details of detection head.

\noindent\textbf{Multi-view Image Acquisition.} To enable an agent to perceive the surroundings through camera, we build a new 3D detection dataset ${\text{Matterport3D}}^2$ on multi-view images captured by camera~\cite{chang2017matterport3d}, which differs from the previous whole-scene detection~\cite{dai2017scannet,song2015sun,tsai2022towards,baruch1arkitscenes} based on point clouds~\cite{qi2017pointnet,qi2017pointnet++}. During navigation, the agent revolves around the direction of gravity to capture the RGB images in $90$ building-scale scenes. The original dataset~\cite{chang2017matterport3d} provides information on the object center and segments throughout the entire scene.

\noindent\textbf{Amodal Perception for Embodied Agent.} Apart from recognizing the semantics and shapes for visible part of the object, the ability to perceive the whole of an occluded object (\ie, amodal perception)~\cite{wilkes1992active,palmer1999vision,yang2019embodied,zhan2022tri} is also significant for navigation. Since occlusion frequently occurs in the indoor scenes, embodied amodal perception aids the agent in comprehending the persistence of scene layouts that objects possess extents and continue to exist even when they are occluded. We consider the occlusion relationship between objects on center visibility criterion, \ie, an object is considered to be visible if its center is not occluded. To determine the visibility of objects in each image from multi-views, we project the object center onto the multi-view image planes and ascertain whether it is located within the camera frustum~\cite{geiger2012we,baruch1arkitscenes}. Specifically, we establish the transformation from 3D world coordinates to pixel coordinates in the image using the intrinsic and extrinsic parameters of the camera. Then, we obtain a group of corresponding objects for the multi-view images (more details are shown in$_{\!}$ Appendix).

\noindent\textbf{3D Oriented Bounding Box Generation.} We spatially register all objects into an egocentric coordinate system at each panoramic viewpoint. To annotate the objects, we utilize a custom algorithm (\textit{cf}. Appendix) which automatically generates 3D oriented bounding boxes (OBB) for 17 categories of indoor objects, as opposed to the axis-aligned bounding box (AABB) annotations with a fixed yaw angle of zero in the original dataset~\cite{chang2017matterport3d}. OBB surrounds the outline of the objects more tightly than AABB, resulting in more accurate route planning for VLN (see Table\!~\ref{table:OBB}). In addition, the amodal detection on ${\text{Matterport3D}}^2$ follows the same train/val/test splits as previous VLN tasks~\cite{AndersonWTB0S0G18,qi2020reverie}.

\noindent\textbf{Bipartite Matching for BEV Detection.} We construct the 3D detection head~\cite{huang2021bevdet,li2022bevdepth,li2022bevformer} upon BEV features $\bm{B}$ on ${\text{Matterport3D}}^2$. A bipartite matching loss~\cite{stewart2016end,carion2020end,zhudeformable,wang2022detr3d} is used to establish a correspondence between the ground-truth and box prediction, which consists of a focal loss~\cite{lin2017focal} for class labels and a $L1$ loss for bounding box regression. We evaluate different BEV methods~\cite{philion2020lift,wang2022detr3d,li2022bevdepth,li2022bevformer} for indoor detection (see Table\!~\ref{table:bevmodels}). Note that BSG is not constrained to any specific BEV model, allowing for seamless integration of advanced BEV frameworks for VLN.

\subsection{Implementation Details}\label{sec:network}
For ease of training, we employ a separate training strategy of the BEV detection and navigation policy networks, as the initial perception module cannot offer a correct feedback (or rewards) to the navigation policy~\cite{yang2019embodied,wijmans2019embodied}. Therefore, BSG utilizes BEV features encoded by BEVFormer~\cite{li2022bevformer}. Following recent VLN practice~\cite{MajumdarSLAPB20,HaoLLCG20,GuhurTCLS21,ChenGSL21}, pretraining and finetuning paradigm is adopted on a basic model~\cite{ChenGTSL22} equipped with BSG. In this section, we will mainly introduce the details of BSG branch and present the detailed results in Table~\ref{table:Component} (see Appendix for more details).

\noindent\textbf{Voxel Sampling.} For view transformation, we introduce the \textit{voxel sampling} here (Eq.\!~\ref{equ:voxelsam}). The default size of BEV queries is $11\times11$ with four reference points (\ie, $Z\!=\!4$) for each query, and the perception ranges are $[-5.0~m, 5.0~m]$ for $x$ and $y$ axes. Considering the practical height of camera and rooms in \cite{chang2017matterport3d}, the predefined height anchors are uniformly sampled from $[-1.0~m, 2.0~m]$ for $z$ axis. The number of neighboring grids for node embedding is $9$ (Eq.\!~\ref{equ:avegrid}).

\noindent\textbf{BSG Architecture.} Following the recent transformer-based methods~\cite{MajumdarSLAPB20,GuhurTCLS21,Hong0QOG21,ChenGSL21,ChenGTSL22,qiao2022hop}, the pretrained LXMERT~\cite{tan2019lxmert} is utilized for initialization. We use $9$, $2$, and $4$ transformer layers in the text encoder and cross-modal encoder (Eq.\!~\ref{equ:crossgraph}\&\ref{equ:crossbev}), respectively. We keep the other parameters consistent with prior works~\cite{tan2019lxmert,ChenGTSL22}. During the finetuning process, the similar structure variants in the cross-modal encoder are adopted as previous studies~\cite{ChenGSL21,ChenGTSL22}. The fused weight $W_f$ in Eq.\!~\ref{equ:globallocal} is set to $0.5$. For the Bivariate Gaussian weight (Eq.\!~\ref{equ:gaussw}), $\bm{\mu}_{x,y}$ is the zero vector, and $\bm{\sigma}_{x,y}$ is the diagonal matrix with diagonal elements of $2$. We set the weight of $0.7$ for OCM and $0.3$ for~\cite{ChenGTSL22}.

\noindent\textbf{Pretraining.} For the R2R~\cite{AndersonWTB0S0G18} and R4R~\cite{jain2019stay}, we adopt the \textit{Masked Language Modeling} (MLM)~\cite{vaswani2017attention,DevlinCLT19} and \textit{Single-step Action Prediction} (SAP)~\cite{Hong0QOG21,ChenGSL21} as auxiliary tasks in the pretraining stage. For REVERIE~\cite{qi2020reverie}, an additional \textit{Object Grounding} (OG)~\cite{lin2021scene,ChenGTSL22} is used for object reasoning. During the pretraining stage, we train the model with a batch size of $32$ for $100$k iterations, using Adam optimizer~\cite{kingma2014adam} with 1e-4 learning rate. Four RTX 3090 GPUs are used for network training, and only one pretraining task is adopted at each mini-batch with the same sampling ratio.

\noindent\textbf{Finetuning.}$_{\!}$ Following$_{\!}$ standard$_{\!}$ protocol$_{\!}$~\cite{ChenGSL21,ChenGTSL22},$_{\!}$ we finetune the pretrained network with a mixture of \textit{teacher-forcing}~\cite{lamb2016professor} and$_{\!}$ \textit{student-forcing}$_{\!}$ on$_{\!}$ different$_{\!}$ VLN$_{\!}$ datasets.$_{\!}$ On$_{\!}$ REVERIE, the OG loss~\cite{lin2021scene,ChenGTSL22} is also employed for finetuning, and a predefined weight $0.20$ is adopted to balance navigation and object grounding. Moreover, we set the learning rate to 1e-5 and batch size to 8 with $25$k iterations.

\noindent\textbf{Inference.} Once trained, the agent is capable of route planning while considering object context and scene layouts (\S\ref{sec:encodebev}). During the testing phase, we update BSG online (\S\ref{sec:BSG}) and predict a fused action score (\S\ref{sec:decision}). The agent is forced to stop if it exceeds the maximum action steps~\cite{AndersonWTB0S0G18}.

\begin{table*}[t]
\centering
        \resizebox{1\textwidth}{!}{
		\setlength\tabcolsep{2.5pt}
		\renewcommand\arraystretch{1.0}
\begin{tabular}{c||cccccc|cccccc|cccccc}
\hline \thickhline
\rowcolor{mygray}
~ &  \multicolumn{18}{c}{REVERIE} \\
\cline{2-19}
\rowcolor{mygray}
~ &  \multicolumn{6}{c|}{\textit{val} \textit{seen}} & \multicolumn{6}{c|}{\textit{val} \textit{unseen}} & \multicolumn{6}{c}{\textit{test} \textit{unseen}} \\

\cline{2-19}
\rowcolor{mygray}

\multirow{-3}{*}{Models} &\small{TL$\downarrow$} &\small{OSR$\uparrow$} &\small{SR$\uparrow$} &\small{SPL$\uparrow$} &\small{RGS$\uparrow$} &\small{RGSPL$\uparrow$} &\small{TL$\downarrow$} &\small{OSR$\uparrow$} &\small{SR$\uparrow$} &\small{SPL$\uparrow$} &\small{RGS$\uparrow$} &\small{RGSPL$\uparrow$} &\small{TL$\downarrow$} &\small{OSR$\uparrow$} &\small{SR$\uparrow$} &\small{SPL$\uparrow$} &\small{RGS$\uparrow$} &\small{RGSPL$\uparrow$}\\
\hline
\hline

RCM~\cite{wang2019reinforced}    &10.70 &29.44 &23.33 &21.82 &16.23 &15.36    &11.98 &14.23 &9.29  &6.97  &4.89  &3.89      &10.60 &11.68 &7.84  &6.67  &3.67  &3.14  \\

FAST-MATTN~\cite{qi2020reverie}  &16.35 &55.17 &50.53 &45.50 &31.97 &29.66    &45.28 &28.20 &14.40 &7.19  &7.84  &4.67      &39.05 &30.63 &19.88 &11.61 &11.28 &6.08  \\
SIA~\cite{lin2021scene}          &13.61 &65.85 &61.91 &57.08 &45.96 &42.65    &41.53 &44.67 &31.53 &16.28 &22.41 &11.56     &48.61 &44.56 &30.80 &14.85 &19.02 &9.20  \\
RecBERT~\cite{Hong0QOG21}        &13.44 &53.90 &51.79 &47.96 &38.23 &35.61    &16.78 &35.02 &30.67 &24.90 &18.77 &15.27     &15.86 &32.91 &29.61 &23.99 &16.50 &13.51 \\
Airbert~\cite{GuhurTCLS21}       &15.16 &48.98 &47.01 &42.34 &32.75 &30.01    &18.71 &34.51 &27.89 &21.88 &18.23 &14.18     &17.91 &34.20 &30.28 &23.61 &16.83 &13.28 \\
HAMT~\cite{ChenGSL21}            &12.79 &47.65 &43.29 &40.19 &27.20 &25.18    &14.08 &36.84 &32.95 &30.20 &18.92 &17.28     &13.62 &33.41 &30.40 &26.67 &14.88 &13.08 \\
HOP~\cite{qiao2022hop}           &13.80 &54.88 &53.76 &47.19 &38.65 &33.85    &16.46 &36.24 &31.78 &26.11 &18.85 &15.73     &16.38 &33.06 &30.17 &24.34 &17.69 &14.34 \\
TD-STP~\cite{zhao2022target}    &$-$ &$-$ &$-$ &$-$ &$-$ &$-$      &$-$ &39.48 &34.88 &27.32 &21.16 &16.56         &$-$ &40.26 &35.89 &27.51 &19.88 &15.40\\
DUET~\cite{ChenGTSL22}           &13.86 &73.86 &71.75 &63.94 &57.41 &51.14    &22.11 &51.07 &46.98 &33.73 &32.15 &23.03     &21.30 &56.91 &52.51 &36.06 &31.88 &22.06 \\ 
LANA~\cite{wang2023lana}         &15.91 &74.28 &71.94 &62.77 &59.02 &50.34    &23.18 &52.97 &48.31 &33.86 &32.86 &22.77     &18.83 &57.20 &51.72 &36.45 &32.95 &22.85\\
\hline
\textbf{Ours}                    &15.26 &\textbf{78.36} &\textbf{76.18} &\textbf{66.69} &\textbf{61.56} &\textbf{54.02}    &24.71 &\textbf{58.05} &\textbf{52.12} &\textbf{35.59} &\textbf{35.36} &\textbf{24.24}     &22.90 &\textbf{62.83} &\textbf{56.45} &\textbf{38.70} &\textbf{33.15} &\textbf{22.34} \\
\hline
\end{tabular}
}
	\vspace*{-3pt}
\captionsetup{font=small}
	\caption{\small{Quantitative comparison results on REVERIE~\cite{qi2020reverie}. `$-$': unavailable statistics. See \S\ref{sec:VLN} for more details.}}
    \label{table:REVERIE}
\vspace*{-2pt}
\end{table*}

\section{Experiment}\label{sec:experiment}
We first provide the results on VLN benchmarks (\S\ref{sec:VLN}). To verify efficacy of core model designs, we conduct a set of diagnostic studies (\S\ref{ablation}). For comprehensive analysis, we investigate the impact of BEV perception on VLN (\S\ref{bevperception}).

\subsection{Performance on VLN}\label{sec:VLN}

\noindent\textbf{Datasets.}~The experiments are conducted on three datasets. REVERIE~\cite{qi2020reverie} contains high-level instructions describing target locations and objects, with a focus on grounding remote target objects. R2R~\cite{AndersonWTB0S0G18} contains $7,\!189$ shortest-path trajectories, each associated with three step-by-step instructions. The dataset is split into \textit{train}, \textit{val} \textit{seen}, \textit{val} \textit{unseen}, and \textit{test} \textit{unseen} sets with $61$, $56$, $11$, and $18$ scenes, respectively. R4R~\cite{jain2019stay} is an extended variant of R2R by concatenating two adjacent trajectories with longer instructions.

\noindent\textbf{Evaluation Metric.}~Following the standard setting~\cite{AndersonWTB0S0G18,FriedHCRAMBSKD18,ChenGSL21} of VLN task, we use five metrics for evaluation, \ie, Success Rate (SR), Trajectory Length (TL), Oracle Success Rate (OSR), Success rate weighted by Path Length (SPL), and Navigation Error (NE). Two additional evaluation metrics, Remote Grounding Success rate (RGS) and Remote Grounding Success weighted by Path Length (RGSPL), are used for REVERIE~\cite{qi2020reverie,Hong0QOG21,ChenGTSL22}. For R4R~\cite{jain2019stay,wang2021structured,ChenGSL21}, Coverage weighted by Length Score (CLS), normalized Dynamic Time Warping (nDTW), and Success rate weighted nDTW (SDTW) are adopted (more details in Appendix).

\noindent\textbf{Performance on REVERIE~\cite{qi2020reverie}.} Table\!~\ref{table:REVERIE} compares our model with the recent state-of-the-art VLN models on REVERIE dataset. We find that our model outperforms previous approaches across all the evaluation metrics on the three splits. Notably, on the \textit{val} \textit{unseen} split, our model outperforms the previous best model DUET~\cite{ChenGTSL22} by $\bm{5.14\%}$ on SR, $\bm{1.86\%}$ on SPL and $\bm{3.21\%}$ on RGS. On the more challenging \textit{test} \textit{unseen} split, we improve over the baseline by $\bm{3.94\%}$ on SR, $\bm{2.64\%}$ on SPL, and $\bm{1.27\%}$ on RGS. This demonstrates the effectiveness of our architecture design.

\noindent\textbf{Performance on R2R~\cite{AndersonWTB0S0G18}.} Table\!~\ref{table:R2R} presents the comparison results on R2R dataset. We can find that our approach sets new state-of-the-arts for most metrics. For instance, on \textit{val} \textit{unseen}, our model yields SR and SPL of $\bm{74}$ and $\bm{62}$, respectively, while those for the baseline method~\cite{ChenGTSL22} are $72$ and $60$. Our approach improves the performance of DUET by solid margins on \textit{test} \textit{unseen} (\ie, $69\!\!\rightarrow\!\!\bm{73}$ for SR, $59\!\!\rightarrow\!\!\bm{62}$ for SPL). In addition, it also shows significant performance gains in terms of NE (\ie, $3.65\!\rightarrow\!\bm{3.19}$).

\noindent\textbf{Performance on R4R~\cite{jain2019stay}.} Table\!~\ref{table:R4R} shows results on R4R dataset. Our approach outperforms others in most metrics and leads to a promising gain on SR (\ie, $45\!\rightarrow\!\bm{47}$).

\begin{table}[t]
\centering
        \resizebox{0.49\textwidth}{!}{
		\setlength\tabcolsep{2.5pt}
		\renewcommand\arraystretch{1.0}
\begin{tabular}{c||cccc|cccc}
\hline \thickhline
\rowcolor{mygray}
~ &  \multicolumn{8}{c}{R2R} \\
\cline{2-9}
\rowcolor{mygray}
~ & \multicolumn{4}{c|}{\textit{val} \textit{unseen}} & \multicolumn{4}{c}{\textit{test} \textit{unseen}} \\
\cline{2-9}
\rowcolor{mygray}
\multirow{-3}{*}{Models} &TL$\downarrow$ &NE$\downarrow$ &SR$\uparrow$ &SPL$\uparrow$ &TL$\downarrow$ &NE$\downarrow$ &SR$\uparrow$ &SPL$\uparrow$\\
\hline
\hline
Seq2Seq~\cite{AndersonWTB0S0G18}  &8.39    &7.81 &22 &$-$    &8.13  &7.85 &20 &18 \\
SF~\cite{FriedHCRAMBSKD18}        &$-$     &6.62 &35 &$-$    &14.82 &6.62 &35 &28 \\
EnvDrop~\cite{TanYB19}            &10.70   &5.22 &52 &48     &11.66 &5.23 &51 &47 \\
AuxRN~\cite{zhu2020vision}        &$-$     &5.28 &55 &50     &$-$   &5.15 &55 &51 \\
PREVALENT~\cite{HaoLLCG20}        &10.19   &4.71 &58 &53     &10.51 &5.30 &54 &51 \\
RelGraph~\cite{hong2020language}  &9.99    &4.73 &57 &53     &10.29 &4.75 &55 &52 \\
Active Perception~\cite{wang2020active}   &20.60   &4.36 &58 &40     &21.60 &4.33 &60 &41\\
RecBERT~\cite{Hong0QOG21}         &12.01   &3.93 &63 &57     &12.35 &4.09 &63 &57 \\
HAMT~\cite{ChenGSL21}             &11.46   &2.29 &66 &61     &12.27 &3.93 &65 &60 \\
SOAT~\cite{moudgil2021soat}       &12.15   &4.28 &59 &53     &12.26 &4.49 &58 &53 \\
EGP~\cite{deng2020evolving}       &$-$     &4.83 &56 &44     &$-$   &5.34 &53 &42 \\
GBE~\cite{zhu2021soon}            &$-$     &5.20 &54 &43     &$-$   &5.18 &53 &43 \\
SSM~\cite{wang2021structured}     &20.7    &4.32 &62 &45     &20.4  &4.57 &61 &46 \\
CCC~\cite{wang2022counterfactual} &$-$     &5.20 &50 &46     &$-$   &5.30 &51 &48 \\
HOP~\cite{qiao2022hop}            &12.27   &3.80 &64 &57     &12.68 &3.83 &64 &59 \\
LANA~\cite{wang2023lana}          &12.0 &$-$ &68 &62      &12.6  &$-$ &65 &60\\
TD-STP~\cite{zhao2022target}    &$-$  &3.22 &70 &63      &$-$ &3.73 &67 &61\\
DUET~\cite{ChenGTSL22}            &13.94   &3.31 &72 &60     &14.73 &3.65 &69 &59 \\ \hline
\textbf{Ours}                     &14.90   &\textbf{2.89} &\textbf{74} &\textbf{62}    &14.86 &\textbf{3.19} &\textbf{73} &\textbf{62} \\
\hline
\end{tabular}
}
	\vspace*{-6pt}
\captionsetup{font=small}
	\caption{\small{Quantitative results on R2R~\cite{AndersonWTB0S0G18} (more details in \S\ref{sec:VLN}).}}
    \label{table:R2R}
\vspace*{-1pt}
\end{table}

\begin{table}[t]
\centering
        \resizebox{0.45\textwidth}{!}{
		\setlength\tabcolsep{6.5pt}
		\renewcommand\arraystretch{1.0}
\begin{tabular}{c||ccccc}
\hline \thickhline
\rowcolor{mygray}
~ & \multicolumn{5}{c}{R4R \textit{val} \textit{unseen}}  \\
\cline{2-6}
\rowcolor{mygray}
\multirow{-2}{*}{Models}    &NE$\downarrow$ &SR$\uparrow$ &CLS$\uparrow$ &nDTW$\uparrow$ &SDTW$\uparrow$ \\
\hline
\hline
SF~\cite{FriedHCRAMBSKD18}        &8.47    &24   &30   &$-$   &$-$  \\
RCM~\cite{wang2019reinforced}     &$-$     &29   &35   &30    &13   \\
EGP~\cite{deng2020evolving}       &8.00    &30   &44   &37    &18   \\
SSM~\cite{wang2021structured}     &8.27    &32   &53   &39    &19   \\
RelGraph~\cite{hong2020language}  &7.43    &36   &41   &47    &34   \\
RecBERT~\cite{Hong0QOG21}         &6.67    &44   &51   &45    &30   \\
HAMT~\cite{ChenGSL21}             &6.09    &45   &58   &50    &32   \\
LANA~\cite{wang2023lana}          &$-$      &43  &60  &52 &32\\
\hline
\textbf{Ours}                     &6.12    &\textbf{47}   &59   &\textbf{53}    &\textbf{34}   \\
\hline
\end{tabular}
}
	\vspace*{-4pt}
\captionsetup{font=small}
	\caption{\small{Quantitative results on R4R~\cite{jain2019stay} (more details in \S\ref{sec:VLN}).}}
    \label{table:R4R}
\vspace*{-4pt}
\end{table}

\begin{figure*}[t]
	\begin{center}
		\includegraphics[width=0.99\linewidth]{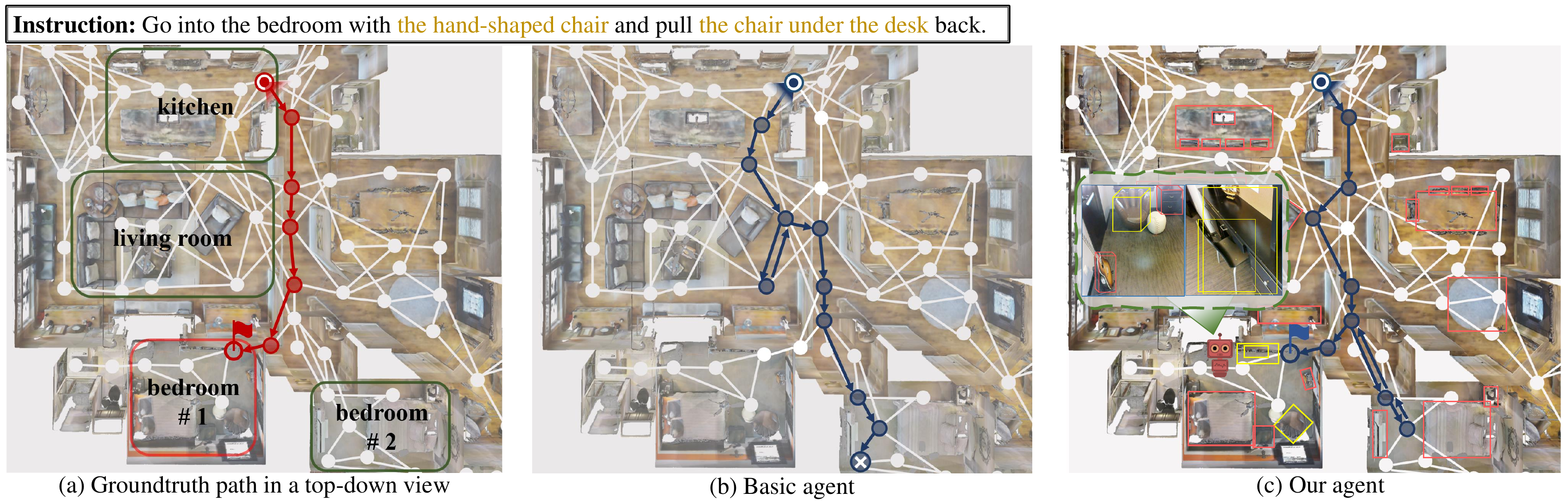}
	\end{center}
	\vspace{-10pt}
	\captionsetup{font=small}
	\caption{\small{A representative visual result on REVERIE dataset~\cite{qi2020reverie} (\S\ref{sec:VLN}). There are two bedrooms and it is difficult to distinguish between them. The basic agent in (b) steps into the bedroom \#2 and ends in failure. With BSG, our agent in (c) returns back to the correction direction and succeeds according to the object context and scene layouts.}}
	\label{fig:reverie}
	\vspace{-10pt}
\end{figure*}

\noindent\textbf{Visual Results.} As shown in Fig.\!~\ref{fig:reverie}, ``bedroom'' is a critical landmark for instruction execution. There are two bedrooms in the environment, which have different objects and geometric context (Fig.\!~\ref{fig:reverie}(a)). However, the basic agent~\cite{ChenGTSL22} navigates a wrong bedroom \#2 and finally fails (Fig.\!~\ref{fig:reverie}(b)). In Fig.\!~\ref{fig:reverie}(c), the BSG enables our agent to perceive the object-aware 3D information, finding ``the chair under the desk'' and ``hand-shaped chair'' to accomplish the task (more visual results are shown in Appendix).

\vspace{-3pt}
\subsection{Diagnostic Experiment}\label{ablation}
To assess the efficacy of essential components of BSG, detailed ablation studies are conducted and the results of \textit{val} \textit{unseen} split of REVERIE~\cite{qi2020reverie} and R2R~\cite{AndersonWTB0S0G18} are shown.

\begin{table}[t]
    \begin{center}
            \resizebox{0.48\textwidth}{!}{
            \setlength\tabcolsep{6.5pt}
            \renewcommand\arraystretch{1.05}
    \begin{tabular}{c|c|ccc|cc}
    \hline \thickhline
    \rowcolor{mygray}
    &  & \multicolumn{3}{c|}{REVERIE} & \multicolumn{2}{c}{R2R} \\
    \cline{3-7}
    \rowcolor{mygray}
    \multicolumn{1}{c|}{\multirow{-2}{*}{\scriptsize{$\#$}}} &\multicolumn{1}{c|}{\multirow{-2}{*}{Models}}         &{SR}$\uparrow$ &SPL$\uparrow$ &{RGS}$\uparrow$ &{SR}$\uparrow$ &SPL$\uparrow$\\
    \hline
    \hline
    1 &Basic agent~\cite{ChenGTSL22}                    &46.98  &33.73  &32.15  &71.52 &60.36  \\ \hline
    2 &BEV Branch                                       &39.03  &25.73  &25.09  &65.56 &52.21  \\
    3 &$\bm {w/o.}$ detection                           &49.25  &32.44  &33.21  &72.65 &60.20  \\
    4 &Full model                                       &\textbf{52.12} &\textbf{35.59} &\textbf{35.36} &\textbf{73.73} &\textbf{62.33}  \\ \hline
    \end{tabular}
    }
    \end{center}
        \vspace*{-13pt}
    \captionsetup{font=small}
	\caption{\small{Ablation study of overall design on \textit{val} \textit{unseen} of REVERIE~\cite{qi2020reverie} and R2R~\cite{AndersonWTB0S0G18}. See \S\ref{ablation} for more details.}}
    \label{table:Component}
    \vspace*{-1pt}
\end{table}

\begin{table}[t]
    \begin{center}
            \resizebox{0.48\textwidth}{!}{
            \setlength\tabcolsep{10pt}
            \renewcommand\arraystretch{1.05}
    \begin{tabular}{l|c|ccc|cc}
    \hline \thickhline
    \rowcolor{mygray}
    & &\multicolumn{3}{c|}{REVERIE} &\multicolumn{2}{c}{R2R}  \\
    \cline{3-7}
    \rowcolor{mygray}
    \multicolumn{1}{c|}{\multirow{-2}{*}{\scriptsize{$\#$}}} &\multicolumn{1}{c|}{\multirow{-2}{*}{$\lvert{{\Omega}^{\rm{n}}}\rvert$}}  &{SR}$\uparrow$ &SPL$\uparrow$ &{RGS}$\uparrow$ &{SR}$\uparrow$ &SPL$\uparrow$ \\
    \hline
    \hline
                               1       &4    &51.33  &34.34  &34.86  &72.89 &62.07  \\
                               2       &9    &\textbf{52.12} &\textbf{35.59} &\textbf{35.36} &\textbf{73.73} &\textbf{62.33}  \\
                               3       &16   &51.71  &34.11  &34.54  &73.26 &61.99  \\
    \hline
    \end{tabular}
    }
    \end{center}
        \vspace*{-14pt}
    \captionsetup{font=small}
	\caption{\small{Ablation study of node embeddings on \textit{val} \textit{unseen} of REVERIE~\cite{qi2020reverie} and R2R~\cite{AndersonWTB0S0G18}. See \S\ref{ablation} for more details.}}
    \label{table:neighborhood}
    \vspace*{-5pt}
\end{table}

\begin{table}[t]
    \begin{center}
            \resizebox{0.48\textwidth}{!}{
            \setlength\tabcolsep{7pt}
            \renewcommand\arraystretch{1.05}
    \begin{tabular}{c|ccc|cc}
    \hline \thickhline
    \rowcolor{mygray}
    &\multicolumn{3}{c|}{REVERIE} &\multicolumn{2}{c}{R2R}  \\
    \cline{2-6}
    \rowcolor{mygray}
    \multicolumn{1}{c|}{\multirow{-2}{*}{Updating}} &{SR}$\uparrow$ &SPL$\uparrow$ &{RGS}$\uparrow$ &{SR}$\uparrow$ &SPL$\uparrow$ \\
    \hline
    \hline
                                  $\bm {w/o.}$ BEV updating      &50.30    &34.05    &35.05    &72.29   &60.77    \\
                                  $\bm {w.}$ BEV updating         &\textbf{52.12} &\textbf{35.59} &\textbf{35.36} &\textbf{73.73} &\textbf{62.33}  \\
    \hline
    \end{tabular}
    }
    \end{center}
        \vspace*{-14pt}
    \captionsetup{font=small}
	\caption{\small{Ablation study of \textit{BEV updating} on \textit{val} \textit{unseen} of REVERIE~\cite{qi2020reverie} and R2R~\cite{AndersonWTB0S0G18}. See \S\ref{ablation} for more details.}}
    \label{table:gridnode}
    \vspace*{-1pt}
\end{table}

\noindent\textbf{Overall Design.} We first investigate the effectiveness of our overall design. The results presented in row \#1, \#2, and \#4 of Table~\ref{table:Component} indicate that adding BEV branch leads to a promising gain over the basic agent~\cite{ChenGTSL22} across all metrics. From row \#3 and \#4, we improve the model by using additional detection loss $\bm{2.87}\%$ on SR of REVERIE, $\bm{3.15}\%$ on RGS of REVERIE, and $\bm{2.13}\%$ on SPL of R2R.

\noindent\textbf{Neighborhood for Node Embeddings.} We next validate the design of node embeddings. For each navigable candidate node, we employ its neighboring grid representations to construct the node embeddings (\textit{cf}. Eq.\!~\ref{equ:avegrid}). In Table\!~\ref{table:neighborhood}, it can be observed that insufficient neighboring grids, as seen in rows \#$1$ and \#$2$, cannot represent the node well for navigation. On the other hand, from row \#$2$ and \#$3$, selecting too many neighboring grids can impact the discriminability of node embeddings due to a large number of overlap between candidate neighborhoods (see Fig.\!~\ref{fig:temporalagg}).

\begin{table}[t]
    \begin{center}
            \resizebox{0.48\textwidth}{!}{
            \setlength\tabcolsep{2.5pt}
            \renewcommand\arraystretch{1.05}
    \begin{tabular}{l|c|cc|ccc|cc}
    \hline \thickhline
    \rowcolor{mygray}
    & & \multicolumn{2}{c|}{Decision Space} & \multicolumn{3}{c|}{REVERIE} & \multicolumn{2}{c}{R2R} \\
    \cline{3-9}
    \rowcolor{mygray}
    \multicolumn{1}{c|}{\multirow{-2}{*}{\scriptsize{$\#$}}} &\multicolumn{1}{c|}{\multirow{-2}{*}{Models}}  &Graph &Grid           &{SR}$\uparrow$ &SPL$\uparrow$ &{RGS}$\uparrow$ &{SR}$\uparrow$ &SPL$\uparrow$\\
    \hline
    \hline
    1 &Basic agent~\cite{ChenGTSL22}                   &$-$           &$-$             &46.98  &33.73  &32.15  &71.52 &60.36  \\ \hline
    2 &\multicolumn{1}{c|}{\multirow{3}{*}{Variants}}  &\checkmark    &                &50.18  &33.94  &33.66  &73.02 &60.76   \\
    3 &                                              &              &\checkmark      &48.25  &34.34  &34.02  &72.79 &61.54   \\
    4* &                                              &              &\checkmark      &51.27  &34.56  &35.20  &73.10 &61.88   \\ \hline
    5 &Full model                                     &\checkmark    &\checkmark      &\textbf{52.12} &\textbf{35.59} &\textbf{35.36} &\textbf{73.73} &\textbf{62.33}  \\ \hline
    \end{tabular}
    }
    \end{center}
        \vspace*{-13pt}
    \captionsetup{font=small}
	\caption{\small{Ablation study of fused decision-making on \textit{val} \textit{unseen} of REVERIE~\cite{qi2020reverie} and R2R~\cite{AndersonWTB0S0G18}. `$*$' denotes using uniform weight instead of \textit{Bivariate} \textit{Gaussian} (Eq. \ref{equ:grid2node}). More details in \S\ref{ablation}.}}
    \label{table:decisionspace}
    \vspace*{-5pt}
\end{table}

\noindent\textbf{BEV Updating Strategy.} At each step, we update BEV features by \textit{cross-attention}, and then use the modified BEV grids to revise node embeddings (\textit{cf}. Eq.\!~\ref{equ:bevupdate}). In Table\!~\ref{table:gridnode}, the variant of model that does not include \textit{BEV updating} leads to inferior performance compared to full model.

\noindent\textbf{Fused Decision-Making.} The results in row \#1, \#2, and \#3 of Table\!~\ref{table:decisionspace} suggest that both graph and grid-level decision space of BSG facilitate the navigation (\textit{cf}. \S\ref{sec:decision}). From row \#4, using \textit{Bivariate} \textit{Gaussian} weights results in better performance compared to assigning uniform weights, as it$_{\!}$ takes$_{\!}$ into account the varying contribution of each BEV$_{\!}$ grid to$_{\!}$ the node$_{\!}$ based$_{\!}$ on the$_{\!}$ relative distances.

\subsection{Analysis on BEV Encoding}\label{bevperception}
In this section, we present the detection results on \textit{val} \textit{unseen} of $\text{Matterport3D}^2$. For evaluation, we utilize mean Average Precision (mAP) and mean Average Recall (mAR), with Intersection over Union (IoU) thresholds of $0.50$, following standard protocols~\cite{lin2014microsoft,caesar2020nuscenes,song2015sun,dai2017scannet}. Then, we provide a quantitative analysis of how BEV detection affects VLN performance, including different types of BEV models (\textit{depth prediction}~\cite{philion2020lift,li2022bevdepth} and \textit{voxel sampling}~\cite{li2022bevformer}) and the ablation study on the superior model~\cite{li2022bevformer}.

\begin{table}[t]
    \begin{center}
            \resizebox{0.48\textwidth}{!}{
            \setlength\tabcolsep{3pt}
            \renewcommand\arraystretch{1.05}
    \begin{tabular}{c||cc||ccc|cc}
    \hline \thickhline
    \rowcolor{mygray}
    &\multicolumn{2}{c||}{\small{$\text{Matterport3D}^2$}} &\multicolumn{3}{c|}{REVERIE} & \multicolumn{2}{c}{R2R} \\
    \cline{4-8}
    \rowcolor{mygray}
    \multicolumn{1}{c||}{\multirow{-2}{*}{BEV Models}}  &mAP$\uparrow$ &mAR$\uparrow$ &{SR}$\uparrow$ &SPL$\uparrow$ &{RGS}$\uparrow$ &{SR}$\uparrow$ &SPL$\uparrow$ \\
    \hline
    \hline
                              LSS~\cite{philion2020lift}            &0.188   &0.270    &$50.83$    &$34.43$   &$33.19$   &$72.25$   &$61.30$ \\

                              BEVDepth~\cite{li2022bevdepth}        &0.252   &0.443  &$51.06$    &$34.35$   &$34.13$   &$72.77$   &$61.38$ \\ \hline
                              BEVFormer~\cite{li2022bevformer}      &\textbf{0.299}  &\textbf{0.488}  &\textbf{52.12}  &\textbf{35.59} &\textbf{35.36} &\textbf{73.73} &\textbf{62.33} \\
    \hline
    \end{tabular}
    }
    \end{center}
        \vspace*{-14pt}
    \captionsetup{font=small}
	\caption{\small{Ablation study of different BEV models on \textit{val} \textit{unseen} of REVERIE~\cite{qi2020reverie} and R2R~\cite{AndersonWTB0S0G18}. See \S\ref{bevperception} for more details.}}
    \label{table:bevmodels}
    \vspace*{-0pt}
\end{table}

\begin{figure}[t]	
\subfigure[mAP on $\text{Matterport3D}^2$]{
\begin{minipage}[t]{0.485\linewidth}
\begin{center}
\includegraphics[width=0.99\linewidth]{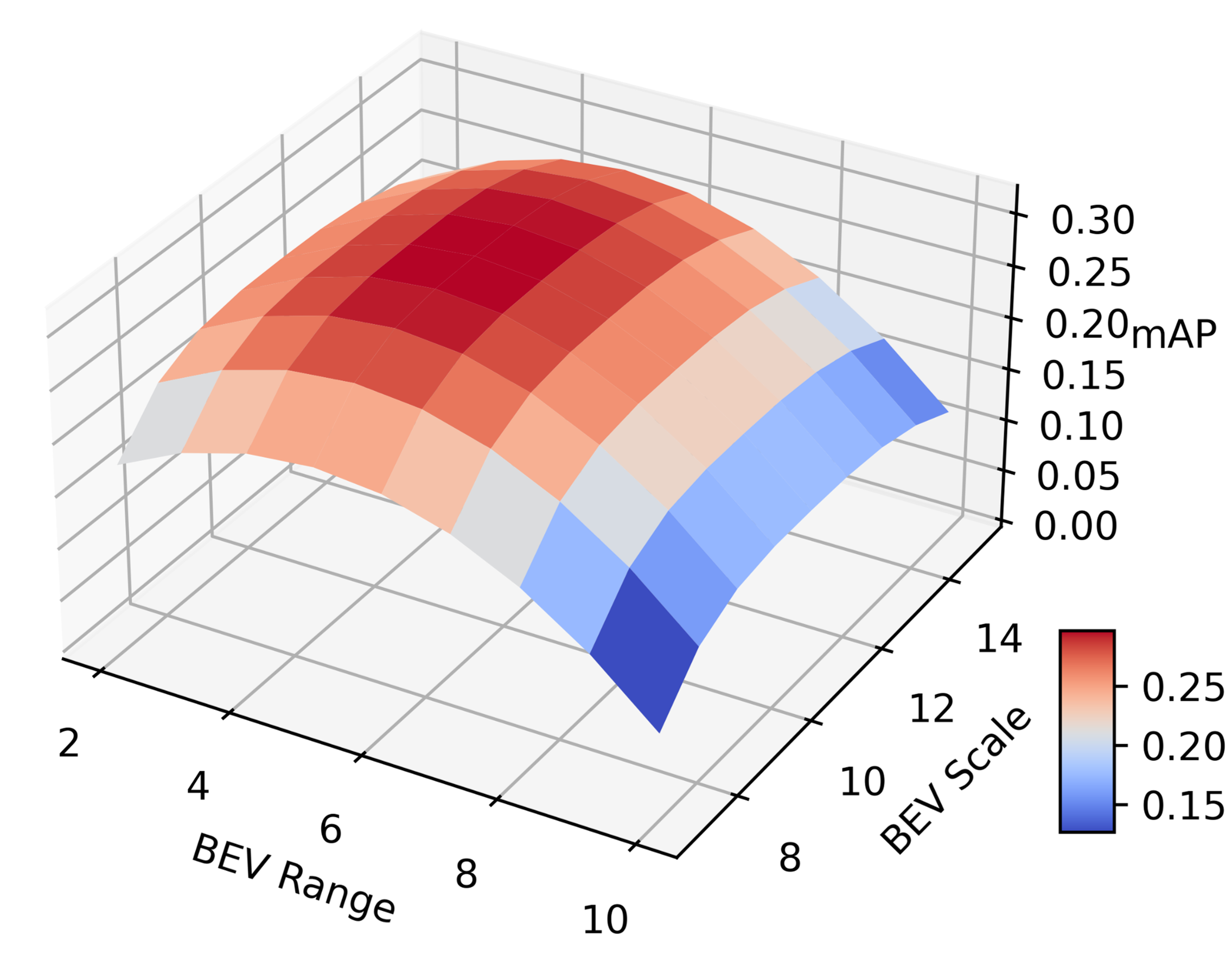}
\end{center}
\end{minipage}%
}
\subfigure[SR on R2R]{
\begin{minipage}[t]{0.485\linewidth}
\begin{center}
\includegraphics[width=0.99\linewidth]{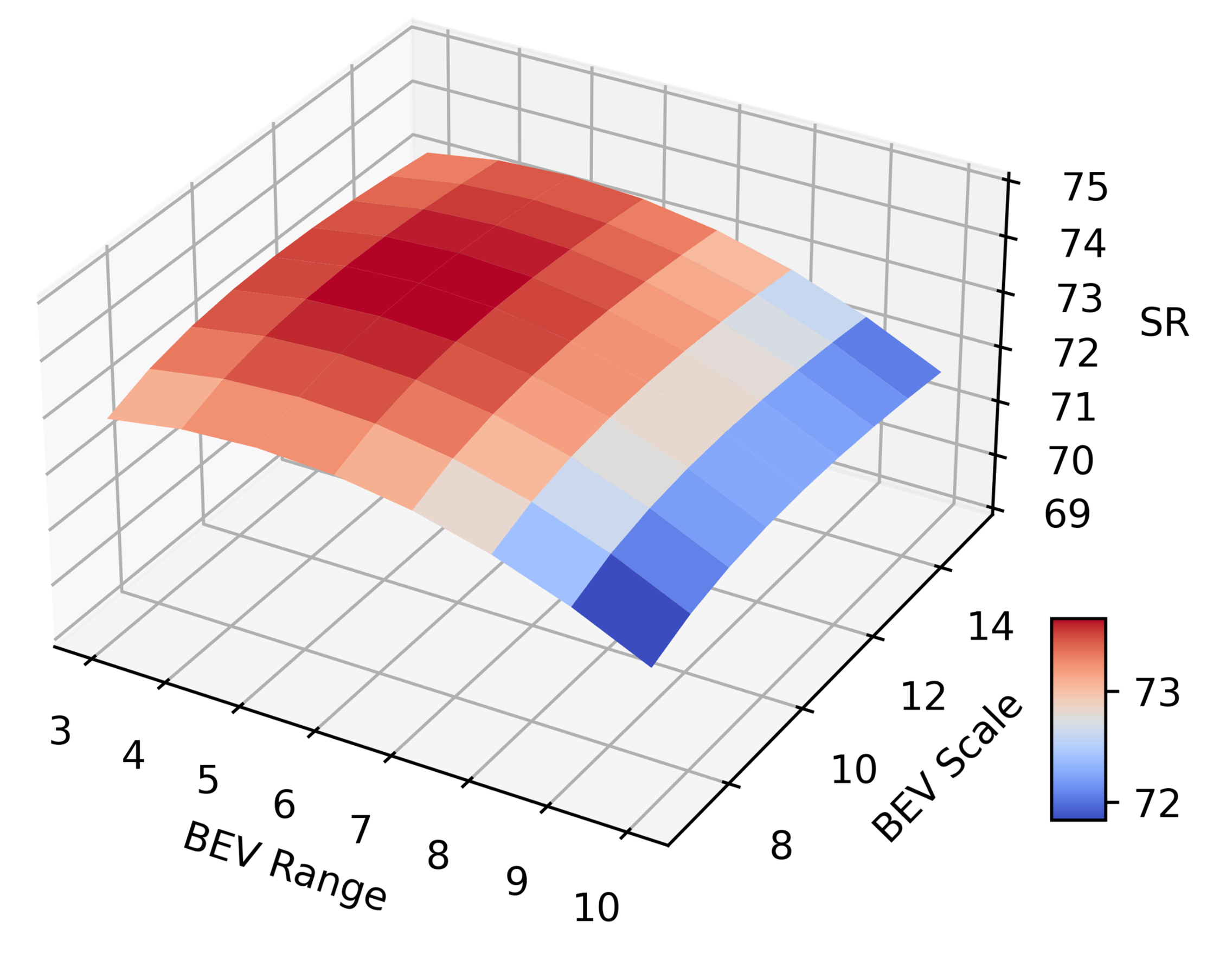}
\end{center}
\end{minipage}%
}%
\vspace{-6pt}
\subfigure[SR on REVERIE]{
\begin{minipage}[t]{0.485\linewidth}
\begin{center}
\includegraphics[width=0.99\linewidth]{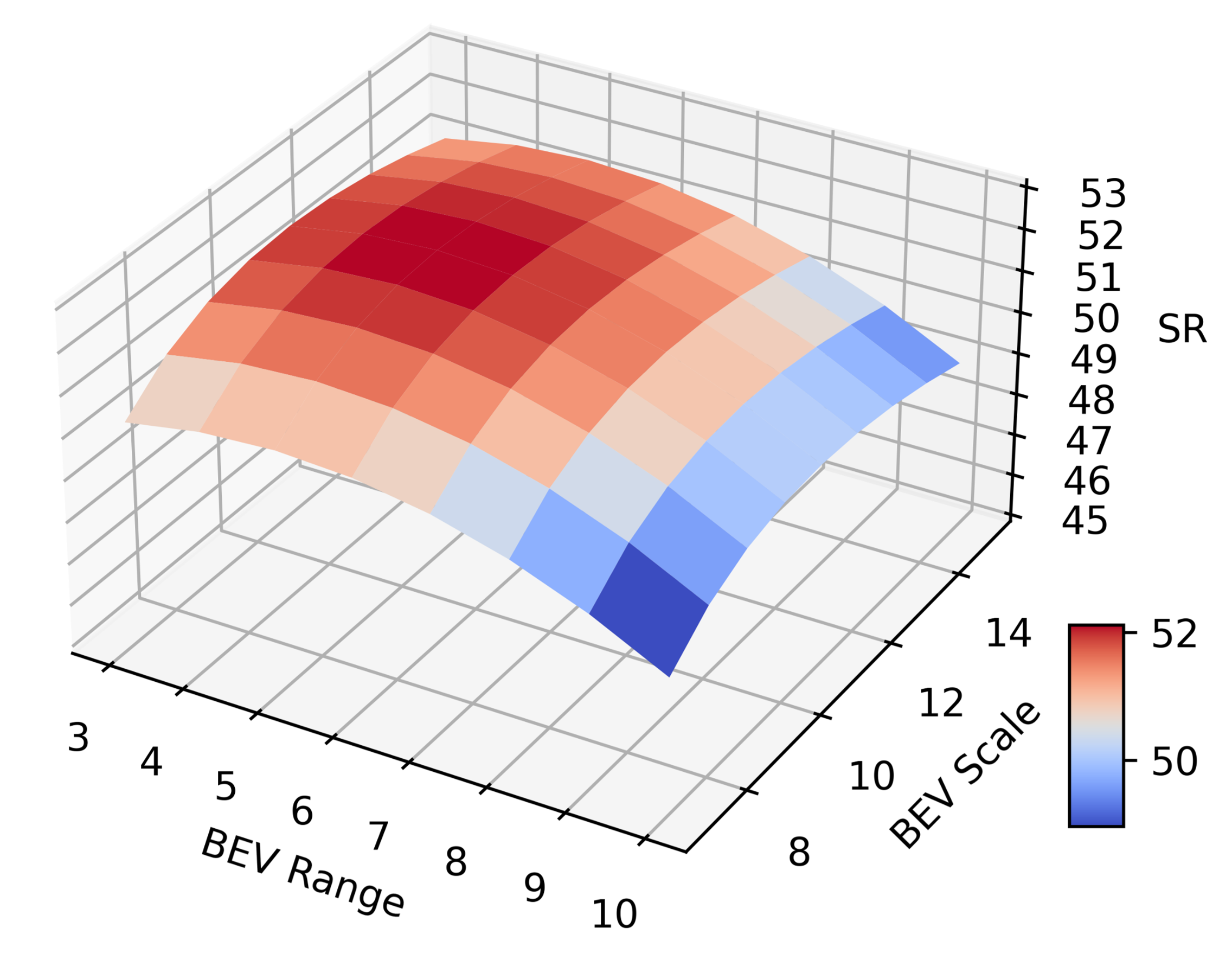}
\end{center}
\end{minipage}%
}%
\subfigure[RGS on REVERIE]{
\begin{minipage}[t]{0.485\linewidth}
\begin{center}
\includegraphics[width=0.99\linewidth]{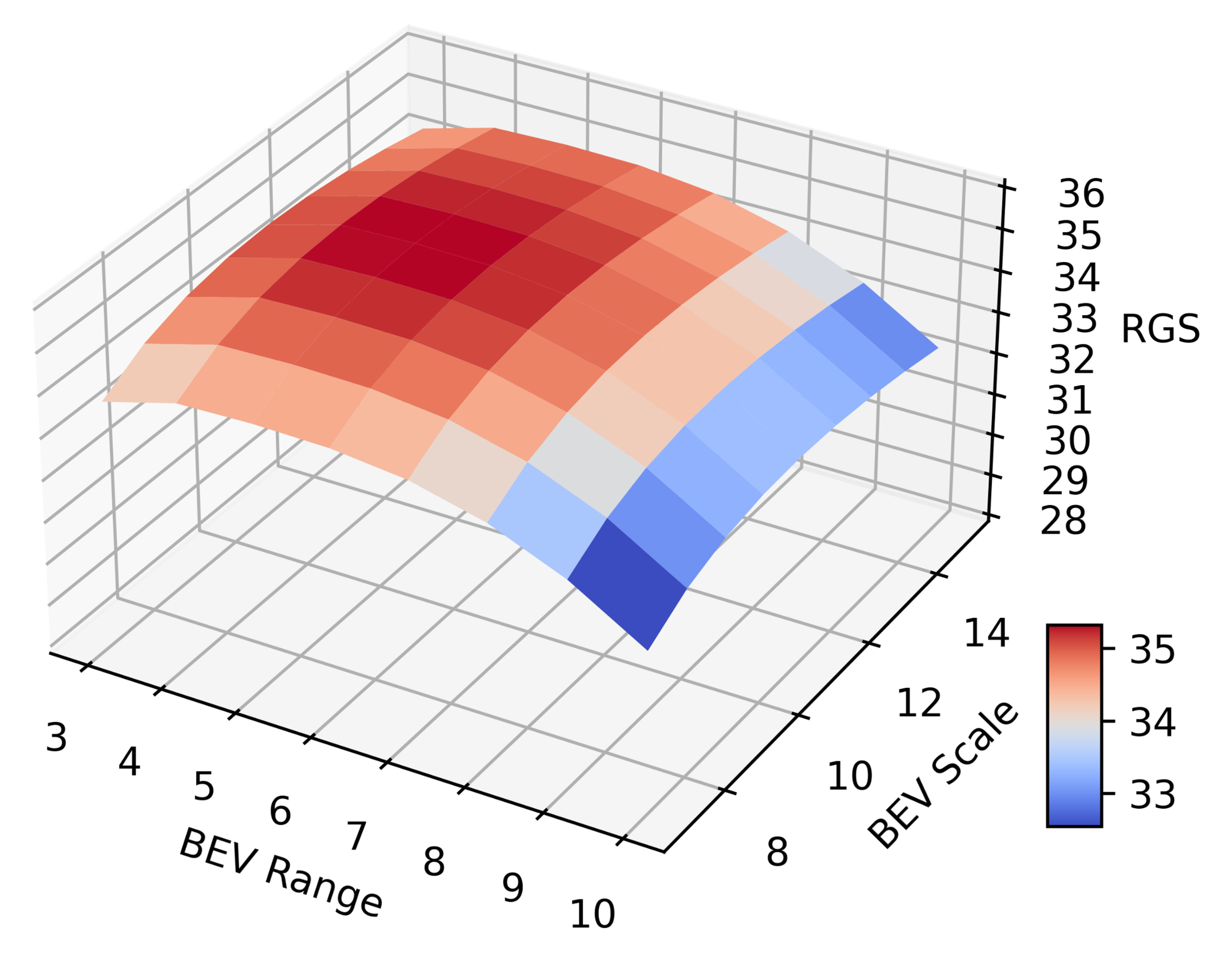}
\end{center}
\end{minipage}%
}%
\vspace{-5pt}
\captionsetup{font=small}
\caption{\small{Ablation study of BEV scale and perception range on \textit{val} \textit{unseen} of REVERIE~\cite{qi2020reverie} and R2R~\cite{AndersonWTB0S0G18} (more details in \S\ref{bevperception})}.}
\label{fig:3dsurface}
	\vspace{-8pt}
\end{figure}

\noindent\textbf{Different$_{\!}$ BEV$_{\!}$ Models.}$_{\!}$ We first compare several representative open-source BEV models~\cite{philion2020lift,li2022bevdepth,li2022bevformer}, which$_{\!}$ are$_{\!}$ divided$_{\!}$ into two$_{\!}$ aspects based$_{\!}$ on$_{\!}$ different$_{\!}$ view transformations. BEVFormer~\cite{li2022bevformer} utilizes voxel sampling to encode 2D features to 3D space (\textit{cf}.~Eq.\!~\ref{equ:voxelsam}), while LSS~\cite{philion2020lift} and BEVDepth~\cite{li2022bevdepth} employ 2D features to predict$_{\!}$  depth information$_{\!}$  and$_{\!}$ then lift  these features to 3D space.$_{\!}$ Note$_{\!}$ that$_{\!}$ BEVDepth~\cite{li2022bevdepth}$_{\!}$ requires explicit depth information$_{\!}$ as additional$_{\!}$ supervision.$_{\!}$ As listed in Table\!~\ref{table:bevmodels}, BEVFormer~\cite{li2022bevformer}$_{\!}$ outperforms$_{\!}$ all$_{\!}$ other$_{\!}$ methods with $\bm{0.299}$ mAP and $\bm{0.488}$ mAR. We adopt BEVFormer~\cite{li2022bevformer} as our BEV baseline.$_{\!}$ Moreover,$_{\!}$ our$_{\!}$ performance$_{\!}$ can$_{\!}$ be further improved with more advanced BEV models.

\noindent\textbf{BEV Scale and Perception Range.} We next delve into the core parameters of our BEV, \ie, scale and perception range (\textit{cf}.~Eq.\!~\ref{equ:voxelsam}). The results are summarized in Fig.\!~\ref{fig:3dsurface}. We find that different scales and perception ranges will affect detection accuracy (\textit{cf}.\! Eq.\!~\ref{equ:voxelsam}). Since$_{\!}$ node$_{\!}$ representations$_{\!}$ are associated with BEV features (\textit{cf}.\! \S\ref{sec:BSG}), better detection performance$_{\!}$ can bring$_{\!}$ more$_{\!}$ gain$_{\!}$ to$_{\!}$ navigation.$_{\!}$ The selection$_{\!}$ of$_{\!}$ an appropriate perception$_{\!}$ range$_{\!}$ should take into account both$_{\!}$ dimensions$_{\!}$ and structure of$_{\!}$ indoor$_{\!}$ environment.

\noindent\textbf{Reference Points.} Table\!~\ref{table:bevpoint} presents a comprehensive analysis of the number of reference points proposed in \S\ref{sec:BSG}. Reference points enable the sampling of multi-view features and their integration into BEV feature (\textit{cf}.~\S\ref{sec:BSG}).

\begin{table}[t]
    \begin{center}
            \resizebox{0.49\textwidth}{!}{
            \setlength\tabcolsep{5pt}
            \renewcommand\arraystretch{1.05}
    \begin{tabular}{l|c||cc||ccc|cc}
    \hline \thickhline
    \rowcolor{mygray}
    & &\multicolumn{2}{c||}{\small{$\text{Matterport3D}^2$}} &\multicolumn{3}{c|}{REVERIE} & \multicolumn{2}{c}{R2R} \\
    \cline{5-9}
    \rowcolor{mygray}
    \multicolumn{1}{c|}{\multirow{-2}{*}{\scriptsize{$\#$}}} &\multicolumn{1}{c||}{\multirow{-2}{*}{$Z$}} &mAP$\uparrow$ &mAR$\uparrow$ &{SR}$\uparrow$ &SPL$\uparrow$ &{RGS}$\uparrow$ &{SR}$\uparrow$ &SPL$\uparrow$ \\
    \hline
    \hline
                              1         &$2$      &0.260  &0.443  &$51.49$    &$35.07$   &$\textbf{36.27}$   &$72.81$   &$60.50$ \\
                              2         &$4$      &\textbf{0.299}  &\textbf{0.488}  &\textbf{52.12}  &\textbf{35.59} &35.36 &\textbf{73.73} &\textbf{62.33} \\
                              3         &$8$      &0.266  &0.438    &50.98    &33.56   &35.34   &72.30   &60.44 \\
    \hline
    \end{tabular}
    }
    \end{center}
        \vspace*{-16pt}
    \captionsetup{font=small}
	\caption{\small{Ablation study of reference points on \textit{val} \textit{unseen} of REVERIE~\cite{qi2020reverie} and R2R~\cite{AndersonWTB0S0G18}. See \S\ref{bevperception} for more details.}}
    \label{table:bevpoint}
    \vspace*{-2pt}
\end{table}

\noindent\textbf{OBB \textit{vs} AABB for Perception and Navigation.} The oriented bounding box (OBB) is more commonly used in 3D perception of real-world scenarios, such as collision detection~\cite{gregory2005framework,bergen1997efficient} and grasp detection~\cite{huebner2008minimum,redmon2015real,zhou2018fully}, compared to the axis-aligned box (AABB). In Table\!~\ref{table:OBB}, using the OBB, the agent's perception performance is better as it provides accurate orientation and scale information (\textit{cf}. \S\ref{sec:encodebev}), resulting in the improved performance in all navigation tasks.

\begin{table}[t]
    \begin{center}
            \resizebox{0.49\textwidth}{!}{
            \setlength\tabcolsep{3.5pt}
            \renewcommand\arraystretch{1.05}
    \begin{tabular}{c||cc||ccc|cc}
    \hline \thickhline
    \rowcolor{mygray}
    &\multicolumn{2}{c||}{\small{$\text{Matterport3D}^2$}} &\multicolumn{3}{c|}{REVERIE} & \multicolumn{2}{c}{R2R} \\
    \cline{4-8}
    \rowcolor{mygray}
    \multicolumn{1}{c||}{\multirow{-2}{*}{Annotation}}  &mAP$\uparrow$ &mAR$\uparrow$ &{SR}$\uparrow$ &SPL$\uparrow$ &{RGS}$\uparrow$ &{SR}$\uparrow$ &SPL$\uparrow$ \\
    \hline
    \hline
                              AABB               &0.266*  &0.491*  &49.25  &32.44  &34.14  &73  &60 \\
                              OBB                &0.299  &0.488    &\textbf{52.12}  &\textbf{35.59}  &\textbf{35.36}  &\textbf{74}  &\textbf{62} \\
    \hline
    \end{tabular}
    }
    \end{center}
        \vspace*{-16pt}
    \captionsetup{font=small}
	\caption{\small{Ablation study of OBB and AABB on \textit{val} \textit{unseen} of REVERIE~\cite{qi2020reverie} and R2R~\cite{AndersonWTB0S0G18}. `$*$' denotes the detection performance on AABB annotations. See \S\ref{bevperception} for more details.}}
    \label{table:OBB}
    \vspace*{-8pt}
\end{table}

\vspace{-1pt}
\section{Conclusion}
Scene understanding is crucial for intelligent navigation in 3D environments. However, current VLN agents rely solely on panoramic observations, lacking the capacity to preserve$_{\!}$ 3D layouts$_{\!}$ and geometric$_{\!}$ cues, and$_{\!}$ hence$_{\!}$ limiting$_{\!}$ their planning$_{\!}$ ability. In$_{\!}$ this paper, we propose a BEV scene graph (BSG) for 3D perception-based VLN, that enables the agent to perceive the scene and access the object layouts. By fusing BSG-based action score and BEV grid-level action score, our approach achieves promising results. This highlights the great potential of BEV perception in VLN.

\newpage


\makesupptitle{Bird’s-Eye-View Scene Graph for Vision-Language Navigation \\\textit{Supplementary Material}}

{
{
\it
	This document provides more details of our approach and additional experimental results, which are organized as follows:
    \begin{itemize}
    \setlength{\itemsep}{0pt}
        \item Model details (\S\ref{sec:model})
        \item Experimental setups (\S\ref{sec:setup})
        \item Additional results and visualization (\S\ref{sec:visualization})
		\item Additional analysis of ${\text{Matterport3D}^2}$ (\S\ref{sec:dataset})
        \item Discussion (\S\ref{sec:discussion})
	\end{itemize}
}

\vspace{2ex}
\section{Model Details} \label{sec:model}
In our model, BEV Scene Graph (BSG) is proposed to enable discriminative decision space (\cf \S\ref{sec:decision}) based on BEV feature.
However, to align with the discrete environments present in the VLN simulator~\cite{AndersonWTB0S0G18,qi2020reverie}, it is necessary to convert the action space into nodes (Fig.\! \ref{fig:complementary}).\! Consequently, BSG can serve as a valuable complement to existing works~\cite{deng2020evolving,wang2021structured,ChenGTSL22} that focus on panoramic decision space (\cf \S\ref{sec:complementary}). Specifically, our approach incorporates a panoramic branch~\cite{ChenGTSL22} . We will give more details on how to train this combined model in \S\ref{sec:pretrain}.

\subsection{Different Decision Space} \label{sec:decisionspace}
\noindent\textbf{Low-level Decision Space.} The early research~\cite{AndersonWTB0S0G18} employed a low-level visuomotor control, which constrained the action space to six actions corresponding to left, right, up, down, forward, and stop. Specifically, the forward action means the agent need to move to the closest reachable viewpoint. The left, right, up and down actions are defined to move the camera by $30$ degrees. Nonetheless, such a visuomotor control posed challenges for the agent to follow instructions accurately and required the agent to memorize extensive sequential inputs.

\noindent\textbf{Panoramic Decision Space.} To enable high-level action reasoning, panoramic decision space~\cite{FriedHCRAMBSKD18} involves discretizing panoramic view of the surrounding environment into $36$ view angles ($12$ headings $\times$ $3$ elevations with $30$ degree intervals). At each location, the agent is limited to a few navigable directions that correspond to these panoramic views. Most existing works~\cite{AndersonWTB0S0G18,FriedHCRAMBSKD18,MaLWAKSX19,Hong0QOG21,ChenGSL21,ChenGTSL22} adopt this decision space. However, due to the adjacency rule (Fig.\!~\ref{fig:introduce}(a)), multiple candidate nodes may correspond to the same panoramic view, resulting in ambiguity during route planning.

\noindent\textbf{BEV Grid Decision Space.} To address the aforementioned constraints, we introduce a grid-level decision space from bird's eye view. Each candidate node corresponds to specific BEV grids (Fig. \ref{fig:introduce}(b)). The node embedding is represented by its neighboring grid features (Fig. \ref{fig:complementary}).

\begin{figure}[t]
	\begin{center}
		\includegraphics[width=0.95\linewidth]{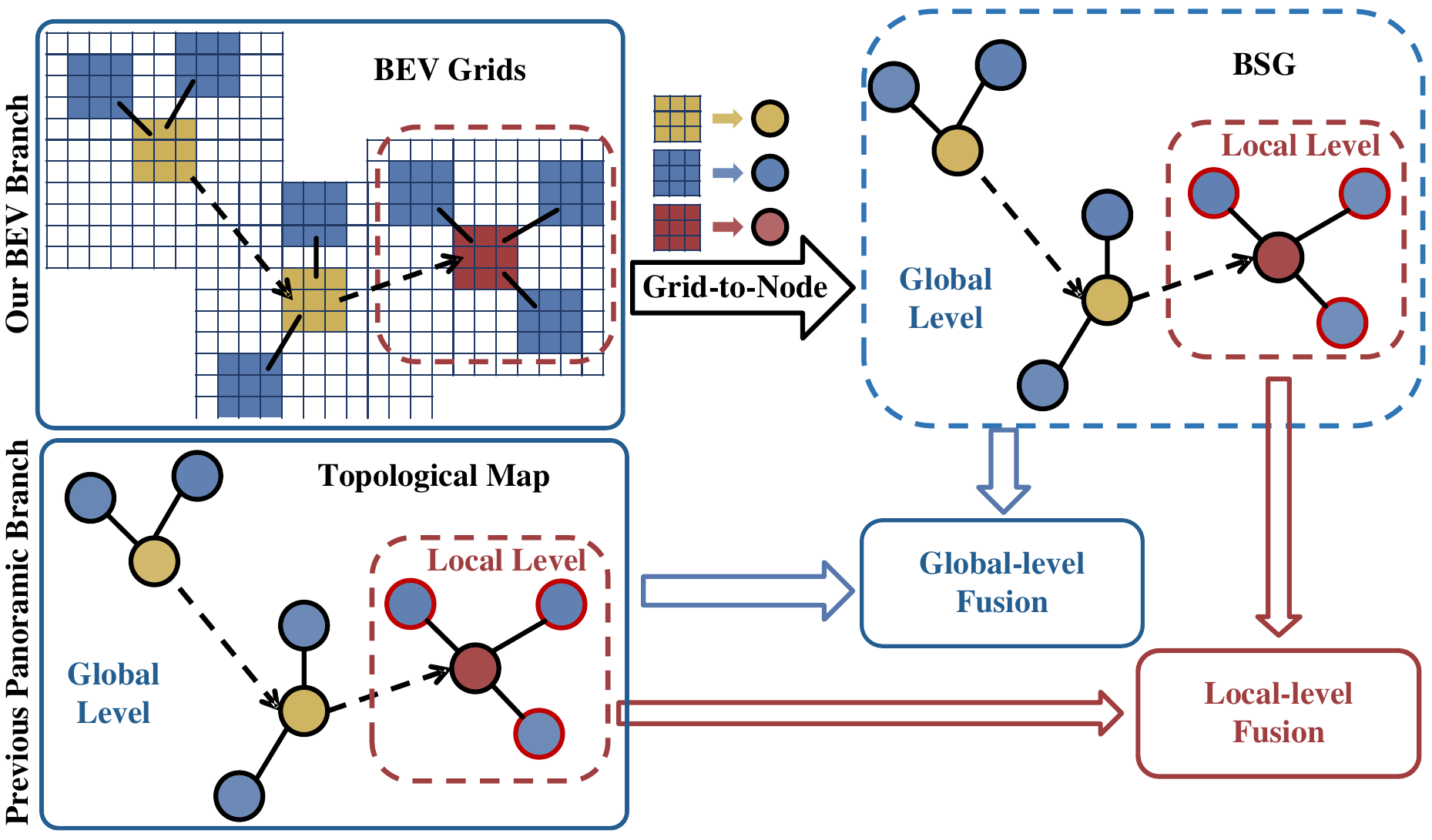}
	\end{center}
	\vspace{-16pt}
	\captionsetup{font=small}
	\caption{\small{Integrating our framework with previous approaches.}}
	\label{fig:complementary}
	\vspace{-14pt}
\end{figure}

\subsection{Complementary to Existing Methods} \label{sec:complementary}
As shown in Fig. \ref{fig:complementary}, our method predicts the next step action by fusing both global and fine-scale local decision-making strategies (see \S\ref{sec:decision}). Specifically, for the topological level, our model predicts the global score on all the navigable nodes, including previously visited and observed nodes, which are similar to previous works~\cite{wang2021structured,deng2020evolving,ChenGTSL22,an2022bevbert}. Meanwhile, for the local level, the local score are for all navigable nodes of the current node, but our model first predicts the BEV grid-level score in the local level then converts to the score of navigable nodes to making a more accurate prediction. Thus, our model can be easily combined with existing work based on panoramic features as shown in Fig. \ref{fig:complementary}. In this paper, we explore the complementary nature of our model with a recent state-of-the-art method~\cite{ChenGTSL22}, which also predicts the global and local score at each step.

\subsection{Detailed Network Architecture on REVERIE}
\noindent\textbf{Object Prediction.} For REVERIE~\cite{qi2020reverie}, an agent is required to identify an object at each step where additional candidate object annotations are provided. To enable fine-grained perception, we incorporate an object prediction module into local branch. Specifically, we adopt the ViT-B/16 pretrained on ImageNet to extract the features of $M$ objects at $t$-th step $O_t\!=\!\{o_m|{o_m}\!\in\!{\mathbb{R}}^{768}\}_{m=1}^M$, and add orientation feature~\cite{MaLWAKSX19,ChenGTSL22} with $\sin$ and $\cos$ values for heading and elevation angles. Then these object features are concatenated with BEV features as visual features, and we adopt a cross-modal transformer on visual and textual features to obtain contextual representations. Finally, grid-level decision score and object score are predicted by \rm{FFN}.

\subsection{Pretraining Objectives} \label{sec:pretrain}
For R2R~\cite{AndersonWTB0S0G18} and R4R~\cite{jain2019stay}, we adopt Masked Language Modeling (MLM)~\cite{vaswani2017attention,DevlinCLT19}, Masked Region Classification (MRC)~\cite{lu2019vilbert,MajumdarSLAPB20,li2020oscar,HaoLLCG20}, and Single-step Action Prediction with Progress Monitoring (SAP-PM)~\cite{MaWAXK19,MaLWAKSX19,Hong0QOG21,ChenGSL21} as auxiliary tasks in the pretraining stage. For REVERIE~\cite{qi2020reverie}, an additional Object Grounding (OG)~\cite{lin2021scene,ChenGTSL22} are used for object reasoning and grounding, and the sample ratio is MLM:MRC:SAP-PM:OG=1:1:1:1. All the auxiliary tasks are based on the input pair $(\mathcal{X},\mathcal{G}_t, \mathcal{T}_t)$, where $\mathcal{X}$ is the textual embedding, $\mathcal{G}_t$ is BSG built at time step $t$, and $\mathcal{T}_t$ is topological map of complementary method~\cite{ChenGTSL22} with panoramic visual feature $V_t$ (\cf \S\ref{sec:complementary}).

\noindent\textbf{MLM.} The task aims to learn grounded language representations in VLN task and cross-modal alignment. It masks some percentage of the input tokens at random, and then predicts those masked tokens based on contextual words and ~\cite{DevlinCLT19}. We randomly mask out one of the word tokens in $\mathcal{X}$ with the probability of $15\%$~\cite{ChenGSL21,ChenGTSL22}, and the final hidden representations corresponding to the [\textit{mask}] token are fed into an output softmax over the instruction vocabulary:
\vspace{-1pt}
\begin{equation}
\small
\begin{aligned}
\mathcal{L}_{\rm {MLM}}=-\log{p({x_i}|\mathcal{X}_{\small{\backslash} i},\mathcal{G}_t,\mathcal{T}_t)},
\end{aligned}
\vspace{-1pt}
\label{equ:MLM}
\end{equation}
where $x_i$ is the textual embedding of the masked token, $\mathcal{X}_{\small{\backslash} i}$ is the masked instruction. We average output embedding of two textual encoders of panoramic branch and BEV branch, and minimize the negative log-likelihood of original words.

\noindent\textbf{MRC.} This task predicts the semantic labels of masked observation features given  instructions and neighboring observations~\cite{ChenGSL21}. We only use this task for panoramic branch, and keep other settings consistent with~\cite{ChenGSL21,ChenGTSL22}.

\noindent\textbf{SAP-PM.} We employs imitation learning to predict the next action~\cite{HaoLLCG20,ChenGSL21,ChenGTSL22}. Specifically, we sample a map-action pair $(\mathcal{G}_t,\mathcal{T}_t,\mathcal{A}_t)$ from the groundtruth trajectory at the $t$-th step, and then the loss of panoramic branch is as follows:
\vspace{-1pt}
\begin{equation}
\footnotesize
\begin{aligned}
\mathcal{L}_{\rm {\scriptscriptstyle SAP}}=\sum\nolimits_{t=1}^{T}{-\log{p(a_t|\mathcal{X},\mathcal{T}_t)}}.
\end{aligned}
\vspace{-1pt}
\label{equ:SAP}
\end{equation}
For our BEV branch, we employ an additional progress monitoring task~\cite{MaWAXK19,MaLWAKSX19} to reflect the navigation progress:
\vspace{-1pt}
\begin{equation}
\footnotesize
\begin{aligned}
\mathcal{L}_{\rm {\scriptscriptstyle SAP-PM}}=\sum\nolimits_{t=1}^{T}{-\log{p(a_t|\mathcal{X},\mathcal{G}_t)}+(y^{pm}_t-p^{pm}_t)^2},
\end{aligned}
\vspace{-1pt}
\label{equ:SAPPM}
\end{equation}
where $y^{pm}_t$ is the normalized distance of length from the current location to the goal as in Eq.(12). We use a weight of $0.5$ to balance $\mathcal{L}_{\rm {\scriptscriptstyle SAP}}$ and $\mathcal{L}_{\rm {\scriptscriptstyle SAP-PM}}$.

\noindent\textbf{OG.} The goal of this task is to predict the best matching object among a set of candidate objects at the current viewpoint~\cite{lin2021scene,ChenGTSL22}. The loss is as follows:
\vspace{-1pt}
\begin{equation}
\footnotesize
\begin{aligned}
\mathcal{L}_{\rm {\scriptscriptstyle OG}}=-\log{p(o_i|\mathcal{X},\mathcal{G}_t,\mathcal{T}_t)},
\end{aligned}
\vspace{-1pt}
\label{equ:OG}
\end{equation}
where $o_i$ is the groundtruth object, and we average the matching score of panoramic branch and BEV branch.

\subsection{Finetuning Objectives} \label{sec:finetune}
Since reinforcement learning reward makes the agent pay more attention on shortest paths rather than path fidelity with instruction~\cite{ChenGSL21}, we alternatively use Teacher-Forcing (TF) and Student-Forcing (SF) for action prediction as behavior cloning (BC):
\vspace{-1pt}
\begin{equation}
\footnotesize
\begin{aligned}
\mathcal{L}_{\rm {\scriptscriptstyle TF}}&=\sum\nolimits_{t=1}^{T}{-\log{p(a_t|\mathcal{X},\mathcal{G}_t,\mathcal{T}_t)}}, \\
\mathcal{L}_{\rm {\scriptscriptstyle SF}}&=\sum\nolimits_{t=1}^{T}{-\log{p(a^*_t|\mathcal{X},\mathcal{G}^*_t,\mathcal{T}^*_t)}},
\end{aligned}
\vspace{-1pt}
\label{equ:finetune}
\end{equation}
where $\mathcal{G}_t$ and $\mathcal{T}_t$ are maps built online following the expert trajectory, $\mathcal{G}^*_t$ and $\mathcal{T}^*_t$ are following the sampling trajectory, and $a^*_t$ is supervised by the pseudo interactive demonstrator in~\cite{ross2011reduction,ChenGTSL22}. On REVERIE, the OG loss is also employed for finetuning, and we adopt a predefined weight $\alpha=0.20$ to balance them:
\vspace{-1pt}
\begin{equation}
\footnotesize
\begin{aligned}
\mathcal{L}=\alpha\mathcal{L}_{\rm {\scriptscriptstyle TF}}+\mathcal{L}_{\rm {\scriptscriptstyle SF}}+\mathcal{L}_{\rm {\scriptscriptstyle OG}}.
\end{aligned}
\vspace{-1pt}
\label{equ:finetune1}
\end{equation}

\section{Experimental Setups} \label{sec:setup}

\subsection{Evaluation Metrics}
\noindent\textbf{VLN.} Following the standard setting~\cite{AndersonWTB0S0G18,FriedHCRAMBSKD18,ChenGSL21} of R2R, there are several metrics for evaluation: (1) Success Rate (SR) considers the percentage of final positions less than 3 m away from the goal location. (2) Trajectory Length (TL) measures the total length of agent trajectories. (3) Oracle Success Rate (OSR) is the success rate if the agent can stop at the closest point to the goal along its trajectory. (4) Success rate weighted by Path Length (SPL) is a trade-off between SR and TL. (5) Navigation Error (NE) refers to the shortest distance between agent’s final position and the goal location. For REVERIE~\cite{qi2020reverie,Hong0QOG21,ChenGTSL22}, there are two additional metrics. (6) Remote Grounding Success rate (RGS) is the success rate of finding the target object. (7) Remote Grounding Success weighted by Path Length (RGSPL) uses the ratio between the length of the ground-truth path and the agent’s path to normalize RGS. For R4R~\cite{jain2019stay,wang2021structured,ChenGSL21}, three metrics are used for instruction fidelity. (8) Coverage weighted by Length Score (CLS) is the product of the path coverage and length score of the agent’s path with respect to reference path. (9) Normalized Dynamic Time Warping (nDTW) and (10) Success rate weighted normalized Dynamic Time Warping (SDTW) measure the order consistency of agent trajectories.

\subsection{Training Details}
\noindent\textbf{VLN.} During the pretraining stage, we train the combined model with a batch size of $32$ for $100$k iterations. We then finetune the model with the batch size of $8$ for $25$k iterations. On REVERIE~\cite{qi2020reverie}, we select the best epoch by SPL on \textit{val} \textit{unseen}. On R2R and R4R~\cite{AndersonWTB0S0G18,jain2019stay}, the best model is selected according to the sum of SR and SPL on \textit{val} \textit{unseen}. For fair comparison, the same synthesize instructions in~\cite{ChenGTSL22} by a speaker model~\cite{FriedHCRAMBSKD18} are also used for REVERIE.

\noindent\textbf{3D Detection.} For BEVFormer~\cite{li2022bevformer}, a static model without using history BEV features is used for 3D detection. We adopt ViT-B/16~\cite{dosovitskiyimage} pretrained on ImageNet as the backbone. The size of the image features are $1280\times{1024}\times{768}$, and we don't utilize the multi-scale features in previous work~\cite{huang2021bevdet,li2022bevdepth,li2022bevformer}. We train this BEV encoder with detection head~\cite{wang2022detr3d,li2022bevformer} using AdamW with a weight decay of $0.01$ for $500$ epoches, a learning rate of $1\!\times\!10^{-4}$.

For LSS~\cite{philion2020lift} and BEVDepth~\cite{li2022bevdepth}, we use ResNet-50 as the image backbone and the image size is processed to $256\times 704$. We don't adopt image or BEV data augmentations. AdamW is used as an optimizer with a learning rate set to $2\!\times\!10^{-4}$ and batch size set to $48$. All experiments are trained for $24$ epochs.

\section{Additional Results and Visualization} \label{sec:visualization}
\noindent\textbf{VLN.} To compare the differences between the two datasets, we also show an example with the same groundtruth path but different instructions in Fig.\!~\ref{fig:vlnvis2}. It shows that detailed instructions in R2R provide additional information that enables a more accurate navigation strategy.

\noindent\textbf{3D Detection.} Table \ref{table:detectiontestunseen} present the detection results on \textit{test} \textit{unseen} in ${\text{Matterport3D}}^2$. For evaluation, we utilize Average Precision (AP) and Average Recall (AR) with Intersection over Union (IoU) thresholds of $0.25$ and $0.50$, following established protocols~\cite{lin2014microsoft,caesar2020nuscenes,song2015sun,dai2017scannet}.  We find that it has good detection performance on larger objects, such as ``bed'' and `sofa' with ${0.535}$ and ${0.394}$ for AP in Table\!~\ref{table:detectiontestunseen}. However, detecting small objects like `picture' and `plant' presents more difficulty since they are almost flat. The detection performance on $\text{Matterport3D}^2$ can be further improved in the future.

\begin{figure*}[t]
	\begin{center}
		\includegraphics[width=0.95\linewidth]{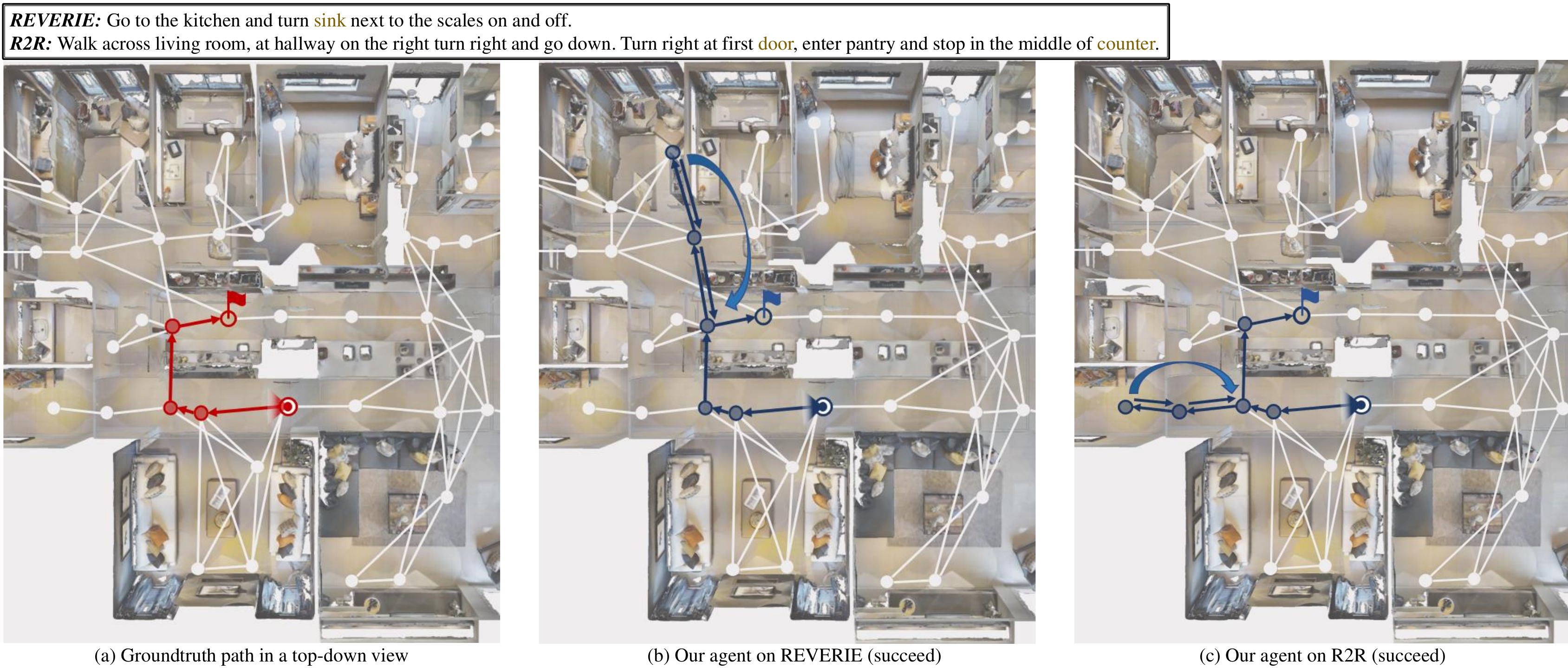}
	\end{center}
	\vspace{-16pt}
	\captionsetup{font=small}
    \vspace{-4pt}
	\caption{\small{Visual results with the same groundtruth path on REVERIE and R2R dataset.}}
	\label{fig:vlnvis2}
	\vspace{-4pt}
\end{figure*}

\section{Additional Analysis of ${\text{Matterport3D}^2}$} \label{sec:dataset}
\subsection{Detailed Annotation Process} \label{sec:annotation}
\noindent\textbf{Images of Skybox from Simulator.} For each panorama in original Matterport3D~\cite{chang2017matterport3d}, the acquisition equipment rotates around the direction of gravity to six distinct orientations, stopping at each to acquire three $1280\times{1024}$ photos from three RGB cameras pointing up, horizontal, and down, respectively. Consequently, each panorama view contains $6\times{3}$ raw images. In the VLN task, most previous works~\cite{FriedHCRAMBSKD18,MaLWAKSX19,qi2020reverie,Hong0QOG21,ChenGTSL22} use the split ``skybox'' images~\cite{chang2017matterport3d} for panoramic viewing. These ``skybox'' images are generated by stitching the raw $6\!\times\!{3}$ images. Then, Matterport3D Simulator~\cite{AndersonWTB0S0G18,FriedHCRAMBSKD18} in the VLN task splits the skybox-based panoramic view into $12\!\times\!{3}$ images with the pre-defined size of $640\times{480}$ (\cf \S\ref{sec:decision}). However, this approach does not produce an explicit view transformation matrix.

\noindent\textbf{Raw Camera Images.} In order to use accurate camera internal and external parameters for projection in 3D detection\footnote{\url{https://github.com/niessner/Matterport/blob/master/data_organization.md}}, we acquire the six raw color images at each viewpoint from the horizontal view for ${\text{Matterport3D}^2}$ dataset. Multi-view perspective images captured by camera can access to the original transformation matrix. Given the camera parameters, the resolution of raw camera image is also fixed. Thus we have to use $1280\!\times\! 1024$ resolution (see \S\ref{sec:BSG}). Specifically, we use the undistorted color images and undistorted camera parameters.

\noindent\textbf{Oriented Bounding Boxes.} Although original dataset~\cite{chang2017matterport3d} provides the axis-aligned bounding boxes, they do not provide accurate annotations for 3D detection. Thus, to conform with standard protocols~\cite{caesar2020nuscenes,sun2020scalability}, we annotate the oriented bounding boxes (OBB) under LiDAR coordinate system~\cite{li2022bevformer,huang2021bevdet}\footnote{\url{https://mmdetection3d.readthedocs.io/en/latest/tutorials/coord_sys_tutorial.html}}, which surrounding the outline of the objects more tightly than the axis-aligned bounding boxes. We apply Principal Component Analysis (PCA) to the $x$ and $y$ coordinates of segments in each object, as each object consists of many annotated segments.

\subsection{Detailed Dataset Statistics}
In Table \ref{table:statistics}, we present the detailed statistics of our ${\text{Matterport3D}^2}$ dataset. At each viewpoint, there are six multi-view images (\cf \S\ref{sec:annotation}). However, since we need to filter the objects at each viewpoint, we only collect the multi-view images of viewpoints that have objects. We use the same \textit{train seen}, \textit{val unseen}, and \textit{test unseen} splits as existing VLN datasets~\cite{AndersonWTB0S0G18,qi2020reverie}.

\begin{table*}
	\centering
	\small
	\resizebox{0.98\textwidth}{!}{
		\setlength\tabcolsep{1.5pt}
		\renewcommand\arraystretch{1.0}
		\begin{tabular}{c|c|ccccccccccccccccc|c}
			\hline\thickhline
			\rowcolor{mygray}
	Split&viewpoints &chair  &door  &table &picture &cabinet &cushion &window &sofa &bed &chest &plant &sink &toilet &monitor &lighting &shelving &appliances &overall\\ \hline\hline
\textit{train seen} &3463 &14665  &18394 &5511  &8493    &3632    &5534    &13918  &1056 &1100&2215  &1875  &1831 &605    &1745    &8171     &2629     &847        &92221 \\
\textit{val unseen}
              &439   &1634   &2456  &863   &1388    &726     &1491    &1501   &176  &97  &211   &223   &179  &48     &72      &762      &380      &107        &12314\\
\textit{test unseen}
              &829   &2388   &4105  &1009  &2492    &1223    &1411    &2365   &289  &285 &277   &1063  &601  &228    &323     &1469     &547      &357        &20432\\ \hline
		\end{tabular}
	}
	\captionsetup{font=small}
    \vspace{-5pt}
	\caption{\small{Statistics of $\text{Matterport3D}^2$ dataset}.}
	\label{table:statistics}
	\vspace{-6pt}
\end{table*}

\section{Discussion} \label{sec:discussion}
\noindent\textbf{Asset License and Consent.}$_{\!}$ In this$_{\!}$ study,$_{\!}$ we$_{\!}$ explore$_{\!}$ vision-language$_{\!}$ navigation using famous datasets, i.e. Matterport3D~\cite{chang2017matterport3d}, R2R~\cite{AndersonWTB0S0G18}, and REVERIE~\cite{qi2020reverie},$_{\!}$ that$_{\!}$ are$_{\!}$ all$_{\!}$ publicly$_{\!}$ available for academic purposes.$_{\!}$ All the code is released under the MIT license.$_{\!}$ We implement all models on the MMDetection3D codebase. MMDetection3D codebase (\url{https://github.com/open-mmlab/mmdetection3d}) is released under Apache 2.0 license.

\noindent\textbf{Broader Impact.} Our work introduces BEV feature for VLN with BSG. Our approach not only achieves a promising improvement of model performance, but also enhances the decision-making by providing grid-level decision score. Furthermore, $\text{Matterport3D}^2$ dataset, which includes oriented bounding boxes for indoor 3D detection, will contribute to future research in the community. It should be noted that our navigation agents are developed and evaluated in virtual simulated environments. Since we primarily trained the model in a static environment where all objects are relatively stationary, deploying the algorithm on a real-world robot may result in collisions with moving objects and cause harm to individuals. Therefore, further research and development should be conducted to ensure safe deployment in real-world scenarios, such as adding more speed sensors to avoid collisions and including additional environments to study potential damage risks.

\begin{table}
	\centering
	\small
	\resizebox{0.4\textwidth}{!}{
		\setlength\tabcolsep{10pt}
		\renewcommand\arraystretch{1.0}
		\begin{tabular}{c||c|c|c|c}
			\hline\thickhline
			\rowcolor{mygray}
			Classes   &$\text{AP}_{25}$  &$\text{AR}_{25}$  &$\text{AP}_{50}$   &$\text{AR}_{50}$  \\ \hline\hline
			cabinet   &0.522   &0.676    &0.348   &0.551  \\
            door      &0.451   &0.649    &0.279   &0.516  \\
            picture   &0.152   &0.334    &0.053   &0.186  \\
            cushion   &0.489   &0.659    &0.281   &0.505  \\
            window    &0.413   &0.570    &0.251   &0.434  \\
            shelving  &0.501   &0.629    &0.320   &0.501  \\
            sofa      &0.663   &0.765    &0.394   &0.581  \\
            lighting  &0.257   &0.486    &0.103   &0.308  \\
            plant     &0.587   &0.729    &0.352   &0.566  \\
            sink      &0.486   &0.654    &0.265   &0.486  \\
            table     &0.487   &0.668    &0.306   &0.525   \\
            bed       &0.691   &0.740    &0.535   &0.649  \\
            toilet    &0.529   &0.645    &0.306   &0.456  \\
            chair     &0.542   &0.695    &0.374   &0.579  \\
            appliances &0.504  &0.613    &0.346   &0.507  \\
            chest     &0.447   &0.607    &0.247   &0.448  \\
            monitor   &0.413   &0.570    &0.264   &0.446  \\ \hline
			Overall   &0.478   &0.629    &0.295   &0.485  \\ \hline
		\end{tabular}
	}
	\captionsetup{font=small}
    \vspace{-5pt}
	\caption{\small{Results on $\text{Matterport3D}^2$ \textit{test} \textit{unseen}.}}
	\label{table:detectiontestunseen}
	\vspace{-6pt}
\end{table}

{\small
\bibliographystyle{unsrt}
\bibliography{egbib}

\begin{thebibliography}{100}

\bibitem{AndersonWTB0S0G18}
Peter Anderson, Qi~Wu, Damien Teney, Jake Bruce, Mark Johnson, Niko
  S{\"{u}}nderhauf, Ian~D. Reid, Stephen Gould, and Anton van~den Hengel.
\newblock Vision-and-language navigation: Interpreting visually-grounded
  navigation instructions in real environments.
\newblock In {\em CVPR}, 2018.

\bibitem{chang2017matterport3d}
Angel Chang, Angela Dai, Thomas Funkhouser, Maciej Halber, Matthias Niebner,
  Manolis Savva, Shuran Song, Andy Zeng, and Yinda Zhang.
\newblock Matterport3d: Learning from rgb-d data in indoor environments.
\newblock In {\em 3DV}, 2017.

\bibitem{FriedHCRAMBSKD18}
Daniel Fried, Ronghang Hu, Volkan Cirik, Anna Rohrbach, Jacob Andreas,
  Louis{-}Philippe Morency, Taylor Berg{-}Kirkpatrick, Kate Saenko, Dan Klein,
  and Trevor Darrell.
\newblock Speaker-follower models for vision-and-language navigation.
\newblock In {\em NeurIPS}, 2018.

\bibitem{wang2019reinforced}
Xin Wang, Qiuyuan Huang, Asli Celikyilmaz, Jianfeng Gao, Dinghan Shen,
  Yuan-Fang Wang, William~Yang Wang, and Lei Zhang.
\newblock Reinforced cross-modal matching and self-supervised imitation
  learning for vision-language navigation.
\newblock In {\em CVPR}, 2019.

\bibitem{TanYB19}
Hao Tan, Licheng Yu, and Mohit Bansal.
\newblock Learning to navigate unseen environments: Back translation with
  environmental dropout.
\newblock In {\em NAACL}, 2019.

\bibitem{hong2020language}
Yicong Hong, Cristian Rodriguez, Yuankai Qi, Qi~Wu, and Stephen Gould.
\newblock Language and visual entity relationship graph for agent navigation.
\newblock In {\em NeurIPS}, 2020.

\bibitem{ZhuQNSBWWEW22}
Wanrong Zhu, Yuankai Qi, Pradyumna Narayana, Kazoo Sone, Sugato Basu, Xin Wang,
  Qi~Wu, Miguel~P. Eckstein, and William~Yang Wang.
\newblock Diagnosing vision-and-language navigation: What really matters.
\newblock In {\em NAACL}, 2022.

\bibitem{lin2022adapt}
Bingqian Lin, Yi~Zhu, Zicong Chen, Xiwen Liang, Jianzhuang Liu, and Xiaodan
  Liang.
\newblock Adapt: Vision-language navigation with modality-aligned action
  prompts.
\newblock In {\em CVPR}, 2022.

\bibitem{MaLWAKSX19}
Chih{-}Yao Ma, Jiasen Lu, Zuxuan Wu, Ghassan AlRegib, Zsolt Kira, Richard
  Socher, and Caiming Xiong.
\newblock Self-monitoring navigation agent via auxiliary progress estimation.
\newblock In {\em ICLR}, 2019.

\bibitem{anderson2019chasing}
Peter Anderson, Ayush Shrivastava, Devi Parikh, Dhruv Batra, and Stefan Lee.
\newblock Chasing ghosts: Instruction following as bayesian state tracking.
\newblock In {\em NeurIPS}, 2019.

\bibitem{ke2019tactical}
Liyiming Ke, Xiujun Li, Yonatan Bisk, Ari Holtzman, Zhe Gan, Jingjing Liu,
  Jianfeng Gao, Yejin Choi, and Siddhartha Srinivasa.
\newblock Tactical rewind: Self-correction via backtracking in
  vision-and-language navigation.
\newblock In {\em CVPR}, 2019.

\bibitem{deng2020evolving}
Zhiwei Deng, Karthik Narasimhan, and Olga Russakovsky.
\newblock Evolving graphical planner: Contextual global planning for
  vision-and-language navigation.
\newblock In {\em NeurIPS}, 2020.

\bibitem{chen2021topological}
Kevin Chen, Junshen~K Chen, Jo~Chuang, Marynel V{\'a}zquez, and Silvio
  Savarese.
\newblock Topological planning with transformers for vision-and-language
  navigation.
\newblock In {\em CVPR}, 2021.

\bibitem{MajumdarSLAPB20}
Arjun Majumdar, Ayush Shrivastava, Stefan Lee, Peter Anderson, Devi Parikh, and
  Dhruv Batra.
\newblock Improving vision-and-language navigation with image-text pairs from
  the web.
\newblock In {\em ECCV}, 2020.

\bibitem{HaoLLCG20}
Weituo Hao, Chunyuan Li, Xiujun Li, Lawrence Carin, and Jianfeng Gao.
\newblock Towards learning a generic agent for vision-and-language navigation
  via pre-training.
\newblock In {\em CVPR}, 2020.

\bibitem{zhu2020vision}
Fengda Zhu, Yi~Zhu, Xiaojun Chang, and Xiaodan Liang.
\newblock Vision-language navigation with self-supervised auxiliary reasoning
  tasks.
\newblock In {\em CVPR}, 2020.

\bibitem{ChenGSL21}
Shizhe Chen, Pierre{-}Louis Guhur, Cordelia Schmid, and Ivan Laptev.
\newblock History aware multimodal transformer for vision-and-language
  navigation.
\newblock In {\em NeurIPS}, 2021.

\bibitem{qiao2022hop}
Yanyuan Qiao, Yuankai Qi, Yicong Hong, Zheng Yu, Peng Wang, and Qi~Wu.
\newblock Hop: history-and-order aware pre-training for vision-and-language
  navigation.
\newblock In {\em CVPR}, 2022.

\bibitem{song2015sun}
Shuran Song, Samuel~P Lichtenberg, and Jianxiong Xiao.
\newblock Sun rgb-d: A rgb-d scene understanding benchmark suite.
\newblock In {\em CVPR}, 2015.

\bibitem{dai2017scannet}
Angela Dai, Angel~X. Chang, Manolis Savva, Maciej Halber, Thomas Funkhouser,
  and Matthias Nie{\ss}ner.
\newblock Scannet: Richly-annotated 3d reconstructions of indoor scenes.
\newblock In {\em CVPR}, 2017.

\bibitem{baruch1arkitscenes}
Gilad Baruch, Zhuoyuan Chen, Afshin Dehghan, Yuri Feigin, Peter Fu, Thomas
  Gebauer, Daniel Kurz, Tal Dimry, Brandon Joffe, Arik Schwartz, et~al.
\newblock Arkitscenes: A diverse real-world dataset for 3d indoor scene
  understanding using mobile rgb-d data.
\newblock In {\em NeurIPS}, 2021.

\bibitem{yang2019embodied}
Jianwei Yang, Zhile Ren, Mingze Xu, Xinlei Chen, David~J Crandall, Devi Parikh,
  and Dhruv Batra.
\newblock Embodied amodal recognition: Learning to move to perceive objects.
\newblock In {\em ICCV}, 2019.

\bibitem{wijmans2019embodied}
Erik Wijmans, Samyak Datta, Oleksandr Maksymets, Abhishek Das, Georgia
  Gkioxari, Stefan Lee, Irfan Essa, Devi Parikh, and Dhruv Batra.
\newblock Embodied question answering in photorealistic environments with point
  cloud perception.
\newblock In {\em CVPR}, 2019.

\bibitem{patil2023advances}
Akshay~Gadi Patil, Supriya~Gadi Patil, Manyi Li, Matthew Fisher, Manolis Savva,
  and Hao Zhang.
\newblock Advances in data-driven analysis and synthesis of 3d indoor scenes.
\newblock {\em arXiv preprint arXiv:2304.03188}, 2023.

\bibitem{qi2020reverie}
Yuankai Qi, Qi~Wu, Peter Anderson, Xin Wang, William~Yang Wang, Chunhua Shen,
  and Anton van~den Hengel.
\newblock Reverie: Remote embodied visual referring expression in real indoor
  environments.
\newblock In {\em CVPR}, 2020.

\bibitem{wang2020active}
Hanqing Wang, Wenguan Wang, Tianmin Shu, Wei Liang, and Jianbing Shen.
\newblock Active visual information gathering for vision-language navigation.
\newblock In {\em ECCV}, 2020.

\bibitem{moudgil2021soat}
Abhinav Moudgil, Arjun Majumdar, Harsh Agrawal, Stefan Lee, and Dhruv Batra.
\newblock Soat: A scene-and object-aware transformer for vision-and-language
  navigation.
\newblock In {\em NeurIPS}, 2021.

\bibitem{ChenGTSL22}
Shizhe Chen, Pierre{-}Louis Guhur, Makarand Tapaswi, Cordelia Schmid, and Ivan
  Laptev.
\newblock Think global, act local: Dual-scale graph transformer for
  vision-and-language navigation.
\newblock In {\em CVPR}, 2022.

\bibitem{GuhurTCLS21}
Pierre{-}Louis Guhur, Makarand Tapaswi, Shizhe Chen, Ivan Laptev, and Cordelia
  Schmid.
\newblock Airbert: In-domain pretraining for vision-and-language navigation.
\newblock In {\em ICCV}, 2021.

\bibitem{Hong0QOG21}
Yicong Hong, Qi~Wu, Yuankai Qi, Cristian~Rodriguez Opazo, and Stephen Gould.
\newblock {VLN} {BERT:} {A} recurrent vision-and-language {BERT} for
  navigation.
\newblock In {\em CVPR}, 2021.

\bibitem{li2022delving}
Hongyang Li, Chonghao Sima, Jifeng Dai, Wenhai Wang, Lewei Lu, Huijie Wang,
  Enze Xie, Zhiqi Li, Hanming Deng, Hao Tian, et~al.
\newblock Delving into the devils of bird's-eye-view perception: A review,
  evaluation and recipe.
\newblock {\em arXiv preprint arXiv:2209.05324}, 2022.

\bibitem{ma2022vision}
Yuexin Ma, Tai Wang, Xuyang Bai, Huitong Yang, Yuenan Hou, Yaming Wang,
  Yu~Qiao, Ruigang Yang, Dinesh Manocha, and Xinge Zhu.
\newblock Vision-centric bev perception: A survey.
\newblock {\em arXiv preprint arXiv:2208.02797}, 2022.

\bibitem{huang2021bevdet}
Junjie Huang, Guan Huang, Zheng Zhu, and Dalong Du.
\newblock Bevdet: High-performance multi-camera 3d object detection in
  bird-eye-view.
\newblock {\em arXiv preprint arXiv:2112.11790}, 2021.

\bibitem{li2022bevdepth}
Yinhao Li, Zheng Ge, Guanyi Yu, Jinrong Yang, Zengran Wang, Yukang Shi,
  Jianjian Sun, and Zeming Li.
\newblock Bevdepth: Acquisition of reliable depth for multi-view 3d object
  detection.
\newblock In {\em AAAI}, 2023.

\bibitem{li2022bevformer}
Zhiqi Li, Wenhai Wang, Hongyang Li, Enze Xie, Chonghao Sima, Tong Lu, Yu~Qiao,
  and Jifeng Dai.
\newblock Bevformer: Learning bird’s-eye-view representation from
  multi-camera images via spatiotemporal transformers.
\newblock In {\em ECCV}, 2022.

\bibitem{huang2023geometric}
Linyan Huang, Huijie Wang, Jia Zeng, Shengchuan Zhang, Liujuan Cao, Rongrong
  Ji, Junchi Yan, and Hongyang Li.
\newblock Geometric-aware pretraining for vision-centric 3d object detection.
\newblock {\em arXiv preprint arXiv:2304.03105}, 2023.

\bibitem{zhang2021end}
Zhejun Zhang, Alexander Liniger, Dengxin Dai, Fisher Yu, and Luc Van~Gool.
\newblock End-to-end urban driving by imitating a reinforcement learning coach.
\newblock In {\em ICCV}, 2021.

\bibitem{hu2022stp3}
Shengchao Hu, Li~Chen, Penghao Wu, Hongyang Li, Junchi Yan, and Dacheng Tao.
\newblock St-p3: End-to-end vision-based autonomous driving via
  spatial-temporal feature learning.
\newblock In {\em ECCV}, 2022.

\bibitem{zhao2022jperceiver}
Haimei Zhao, Jing Zhang, Sen Zhang, and Dacheng Tao.
\newblock Jperceiver: Joint perception network for depth, pose and layout
  estimation in driving scenes.
\newblock In {\em ECCV}, 2022.

\bibitem{hu2022goal}
Yihan Hu, Jiazhi Yang, Li~Chen, Keyu Li, Chonghao Sima, Xizhou Zhu, Siqi Chai,
  Senyao Du, Tianwei Lin, Wenhai Wang, et~al.
\newblock Goal-oriented autonomous driving.
\newblock In {\em CVPR}, 2023.

\bibitem{wang2021structured}
Hanqing Wang, Wenguan Wang, Wei Liang, Caiming Xiong, and Jianbing Shen.
\newblock Structured scene memory for vision-language navigation.
\newblock In {\em CVPR}, 2021.

\bibitem{philion2020lift}
Jonah Philion and Sanja Fidler.
\newblock Lift, splat, shoot: Encoding images from arbitrary camera rigs by
  implicitly unprojecting to 3d.
\newblock In {\em ECCV}, 2020.

\bibitem{reading2021categorical}
Cody Reading, Ali Harakeh, Julia Chae, and Steven~L Waslander.
\newblock Categorical depth distribution network for monocular 3d object
  detection.
\newblock In {\em CVPR}, 2021.

\bibitem{wang2022detr3d}
Yue Wang, Vitor~Campagnolo Guizilini, Tianyuan Zhang, Yilun Wang, Hang Zhao,
  and Justin Solomon.
\newblock Detr3d: 3d object detection from multi-view images via 3d-to-2d
  queries.
\newblock In {\em CoRL}, 2022.

\bibitem{stewart2016end}
Russell Stewart, Mykhaylo Andriluka, and Andrew~Y Ng.
\newblock End-to-end people detection in crowded scenes.
\newblock In {\em CVPR}, 2016.

\bibitem{carion2020end}
Nicolas Carion, Francisco Massa, Gabriel Synnaeve, Nicolas Usunier, Alexander
  Kirillov, and Sergey Zagoruyko.
\newblock End-to-end object detection with transformers.
\newblock In {\em ECCV}, 2020.

\bibitem{elfes1990occupancy}
Alberto Elfes.
\newblock Occupancy grids: A stochastic spatial representation for active robot
  perception.
\newblock In {\em Proceedings of the Sixth Conference on Uncertainty in AI},
  1990.

\bibitem{ChenGG19}
Tao Chen, Saurabh Gupta, and Abhinav Gupta.
\newblock Learning exploration policies for navigation.
\newblock In {\em ICLR}, 2019.

\bibitem{chaplot2020learning}
Devendra~Singh Chaplot, Dhiraj Gandhi, Saurabh Gupta, Abhinav Gupta, and Ruslan
  Salakhutdinov.
\newblock Learning to explore using active neural slam.
\newblock In {\em ICLR}, 2020.

\bibitem{ramakrishnan2021exploration}
Santhosh~K Ramakrishnan, Dinesh Jayaraman, and Kristen Grauman.
\newblock An exploration of embodied visual exploration.
\newblock {\em IJCV}, 129:1616--1649, 2021.

\bibitem{HenriquesV18}
Jo{\~{a}}o~F. Henriques and Andrea Vedaldi.
\newblock Mapnet: An allocentric spatial memory for mapping environments.
\newblock In {\em CVPR}, 2018.

\bibitem{CartillierRJLEB21}
Vincent Cartillier, Zhile Ren, Neha Jain, Stefan Lee, Irfan Essa, and Dhruv
  Batra.
\newblock Semantic mapnet: Building allocentric semantic maps and
  representations from egocentric views.
\newblock In {\em AAAI}, 2021.

\bibitem{georgakis2022cross}
Georgios Georgakis, Karl Schmeckpeper, Karan Wanchoo, Soham Dan, Eleni
  Miltsakaki, Dan Roth, and Kostas Daniilidis.
\newblock Cross-modal map learning for vision and language navigation.
\newblock In {\em CVPR}, 2022.

\bibitem{chen2022weakly}
Peihao Chen, Dongyu Ji, Kunyang Lin, Runhao Zeng, Thomas~H Li, Mingkui Tan, and
  Chuang Gan.
\newblock Weakly-supervised multi-granularity map learning for
  vision-and-language navigation.
\newblock In {\em NeurIPS}, 2022.

\bibitem{an2022bevbert}
Dong An, Yuankai Qi, Yangguang Li, Yan Huang, Liang Wang, Tieniu Tan, and Jing
  Shao.
\newblock Bevbert: Topo-metric map pre-training for language-guided navigation.
\newblock {\em arXiv preprint arXiv:2212.04385}, 2022.

\bibitem{jain2019stay}
Vihan Jain, Gabriel Magalhaes, Alexander Ku, Ashish Vaswani, Eugene Ie, and
  Jason Baldridge.
\newblock Stay on the path: Instruction fidelity in vision-and-language
  navigation.
\newblock In {\em ACL}, 2019.

\bibitem{MaWAXK19}
Chih{-}Yao Ma, Zuxuan Wu, Ghassan AlRegib, Caiming Xiong, and Zsolt Kira.
\newblock The regretful agent: Heuristic-aided navigation through progress
  estimation.
\newblock In {\em CVPR}, 2019.

\bibitem{vasudevan2021talk2nav}
Arun~Balajee Vasudevan, Dengxin Dai, and Luc Van~Gool.
\newblock Talk2nav: Long-range vision-and-language navigation with dual
  attention and spatial memory.
\newblock {\em IJCV}, 129:246--266, 2021.

\bibitem{LiLXBCGSC19}
Xiujun Li, Chunyuan Li, Qiaolin Xia, Yonatan Bisk, Asli Celikyilmaz, Jianfeng
  Gao, Noah~A. Smith, and Yejin Choi.
\newblock Robust navigation with language pretraining and stochastic sampling.
\newblock In {\em EMNLP}, 2019.

\bibitem{DevlinCLT19}
Jacob Devlin, Ming{-}Wei Chang, Kenton Lee, and Kristina Toutanova.
\newblock {BERT:} pre-training of deep bidirectional transformers for language
  understanding.
\newblock In {\em NAACL}, 2019.

\bibitem{PashevichS021}
Alexander Pashevich, Cordelia Schmid, and Chen Sun.
\newblock Episodic transformer for vision-and-language navigation.
\newblock In {\em ICCV}, 2021.

\bibitem{liu2021vision}
Chong Liu, Fengda Zhu, Xiaojun Chang, Xiaodan Liang, Zongyuan Ge, and Yi-Dong
  Shen.
\newblock Vision-language navigation with random environmental mixup.
\newblock In {\em ICCV}, 2021.

\bibitem{li2022envedit}
Jialu Li, Hao Tan, and Mohit Bansal.
\newblock Envedit: Environment editing for vision-and-language navigation.
\newblock In {\em CVPR}, 2022.

\bibitem{koh2021pathdreamer}
Jing~Yu Koh, Honglak Lee, Yinfei Yang, Jason Baldridge, and Peter Anderson.
\newblock Pathdreamer: A world model for indoor navigation.
\newblock In {\em ICCV}, 2021.

\bibitem{fu2020counterfactual}
Tsu-Jui Fu, Xin~Eric Wang, Matthew~F Peterson, Scott~T Grafton, Miguel~P
  Eckstein, and William~Yang Wang.
\newblock Counterfactual vision-and-language navigation via adversarial path
  sampler.
\newblock In {\em ECCV}, 2020.

\bibitem{wang2022counterfactual}
Hanqing Wang, Wei Liang, Jianbing Shen, Luc Van~Gool, and Wenguan Wang.
\newblock Counterfactual cycle-consistent learning for instruction following
  and generation in vision-language navigation.
\newblock In {\em CVPR}, 2022.

\bibitem{wang2023lana}
Xiaohan Wang, Wenguan Wang, Jiayi Shao, and Yi~Yang.
\newblock Lana: A language-capable navigator for instruction following and
  generation.
\newblock In {\em CVPR}, 2023.

\bibitem{huang2019transferable}
Haoshuo Huang, Vihan Jain, Harsh Mehta, Alexander Ku, Gabriel Magalhaes, Jason
  Baldridge, and Eugene Ie.
\newblock Transferable representation learning in vision-and-language
  navigation.
\newblock In {\em ICCV}, 2019.

\bibitem{AnQHWWT21}
Dong An, Yuankai Qi, Yan Huang, Qi~Wu, Liang Wang, and Tieniu Tan.
\newblock Neighbor-view enhanced model for vision and language navigation.
\newblock In {\em ACM MM}, 2021.

\bibitem{zhu2021soon}
Fengda Zhu, Xiwen Liang, Yi~Zhu, Qizhi Yu, Xiaojun Chang, and Xiaodan Liang.
\newblock Soon: Scenario oriented object navigation with graph-based
  exploration.
\newblock In {\em CVPR}, 2021.

\bibitem{zhao2022target}
Yusheng Zhao, Jinyu Chen, Chen Gao, Wenguan Wang, Lirong Yang, Haibing Ren,
  Huaxia Xia, and Si~Liu.
\newblock Target-driven structured transformer planner for vision-language
  navigation.
\newblock In {\em ACM MM}, 2022.

\bibitem{yang2021multiple}
Yi~Yang, Yueting Zhuang, and Yunhe Pan.
\newblock Multiple knowledge representation for big data artificial
  intelligence: framework, applications, and case studies.
\newblock {\em FITEE}, 22(12):1551--1558, 2021.

\bibitem{li2023kerm}
Xiangyang Li, Zihan Wang, Jiahao Yang, Yaowei Wang, and Shuqiang Jiang.
\newblock Kerm: Knowledge enhanced reasoning for vision-and-language
  navigation.
\newblock In {\em CVPR}, 2023.

\bibitem{wang2018look}
Xin Wang, Wenhan Xiong, Hongmin Wang, and William~Yang Wang.
\newblock Look before you leap: Bridging model-free and model-based
  reinforcement learning for planned-ahead vision-and-language navigation.
\newblock In {\em ECCV}, 2018.

\bibitem{ku2020room}
Alexander Ku, Peter Anderson, Roma Patel, Eugene Ie, and Jason Baldridge.
\newblock Room-across-room: Multilingual vision-and-language navigation with
  dense spatiotemporal grounding.
\newblock In {\em EMNLP}, 2020.

\bibitem{krantz2020beyond}
Jacob Krantz, Erik Wijmans, Arjun Majumdar, Dhruv Batra, and Stefan Lee.
\newblock Beyond the nav-graph: Vision-and-language navigation in continuous
  environments.
\newblock In {\em ECCV}, 2020.

\bibitem{xia2018gibson}
Fei Xia, Amir~R Zamir, Zhiyang He, Alexander Sax, Jitendra Malik, and Silvio
  Savarese.
\newblock Gibson env: Real-world perception for embodied agents.
\newblock In {\em CVPR}, 2018.

\bibitem{wang2022towards}
Hanqing Wang, Wei Liang, Luc~V Gool, and Wenguan Wang.
\newblock Towards versatile embodied navigation.
\newblock In {\em NeurIPS}, 2022.

\bibitem{savva2019habitat}
Manolis Savva, Abhishek Kadian, Oleksandr Maksymets, Yili Zhao, Erik Wijmans,
  Bhavana Jain, Julian Straub, Jia Liu, Vladlen Koltun, Jitendra Malik, et~al.
\newblock Habitat: A platform for embodied ai research.
\newblock In {\em ICCV}, 2019.

\bibitem{RamakrishnanGWM21}
Santhosh~Kumar Ramakrishnan, Aaron Gokaslan, Erik Wijmans, Oleksandr Maksymets,
  Alexander Clegg, John Turner, Eric Undersander, Wojciech Galuba, Andrew
  Westbury, Angel~X. Chang, Manolis Savva, Yili Zhao, and Dhruv Batra.
\newblock Habitat-matterport 3d dataset {(HM3D):} 1000 large-scale 3d
  environments for embodied {AI}.
\newblock In {\em NeurIPS Datasets and Benchmarks}, 2021.

\bibitem{szot2021habitat}
Andrew Szot, Alexander Clegg, Eric Undersander, Erik Wijmans, Yili Zhao, John
  Turner, Noah Maestre, Mustafa Mukadam, Devendra~Singh Chaplot, Oleksandr
  Maksymets, et~al.
\newblock Habitat 2.0: Training home assistants to rearrange their habitat.
\newblock In {\em NeurIPS}, 2021.

\bibitem{thrun2002probabilistic}
Sebastian Thrun.
\newblock Probabilistic robotics.
\newblock {\em Communications of the ACM}, 45(3):52--57, 2002.

\bibitem{thrun1998learning}
Sebastian Thrun.
\newblock Learning metric-topological maps for indoor mobile robot navigation.
\newblock {\em Artificial Intelligence}, 99(1):21--71, 1998.

\bibitem{irshad2022semantically}
Muhammad~Zubair Irshad, Niluthpol~Chowdhury Mithun, Zachary Seymour, Han-Pang
  Chiu, Supun Samarasekera, and Rakesh Kumar.
\newblock Semantically-aware spatio-temporal reasoning agent for
  vision-and-language navigation in continuous environments.
\newblock In {\em ICPR}, 2022.

\bibitem{blanco2008toward}
Jose-Luis Blanco, Juan-Antonio Fern{\'a}ndez-Madrigal, and Javier Gonzalez.
\newblock Toward a unified bayesian approach to hybrid metric--topological
  slam.
\newblock {\em IEEE Transactions on Robotics}, 24(2):259--270, 2008.

\bibitem{gomez2020hybrid}
Clara Gomez, Marius Fehr, Alex Millane, Alejandra~C Hernandez, Juan Nieto,
  Ramon Barber, and Roland Siegwart.
\newblock Hybrid topological and 3d dense mapping through autonomous
  exploration for large indoor environments.
\newblock In {\em ICRA}, 2020.

\bibitem{niijima2020city}
Shun Niijima, Ryusuke Umeyama, Yoko Sasaki, and Hiroshi Mizoguchi.
\newblock City-scale grid-topological hybrid maps for autonomous mobile robot
  navigation in urban area.
\newblock In {\em IROS}, 2020.

\bibitem{gupta2013perceptual}
Saurabh Gupta, Pablo Arbelaez, and Jitendra Malik.
\newblock Perceptual organization and recognition of indoor scenes from rgb-d
  images.
\newblock In {\em CVPR}, 2013.

\bibitem{koppula2011semantic}
Hema Koppula, Abhishek Anand, Thorsten Joachims, and Ashutosh Saxena.
\newblock Semantic labeling of 3d point clouds for indoor scenes.
\newblock In {\em NeurIPS}, 2011.

\bibitem{mildenhall2020nerf}
B~Mildenhall, PP~Srinivasan, M~Tancik, JT~Barron, R~Ramamoorthi, and R~Ng.
\newblock Nerf: Representing scenes as neural radiance fields for view
  synthesis.
\newblock In {\em ECCV}, 2020.

\bibitem{jiang2020local}
Chiyu Jiang, Avneesh Sud, Ameesh Makadia, Jingwei Huang, Matthias Nie{\ss}ner,
  Thomas Funkhouser, et~al.
\newblock Local implicit grid representations for 3d scenes.
\newblock In {\em CVPR}, 2020.

\bibitem{WuLHZ021}
Aming Wu, Rui Liu, Yahong Han, Linchao Zhu, and Yi~Yang.
\newblock Vector-decomposed disentanglement for domain-invariant object
  detection.
\newblock In {\em ICCV}, 2021.

\bibitem{niemeyer2020differentiable}
Michael Niemeyer, Lars Mescheder, Michael Oechsle, and Andreas Geiger.
\newblock Differentiable volumetric rendering: Learning implicit 3d
  representations without 3d supervision.
\newblock In {\em CVPR}, 2020.

\bibitem{song2017semantic}
Shuran Song, Fisher Yu, Andy Zeng, Angel~X Chang, Manolis Savva, and Thomas
  Funkhouser.
\newblock Semantic scene completion from a single depth image.
\newblock In {\em CVPR}, 2017.

\bibitem{qi2017pointnet}
Charles~R Qi, Hao Su, Kaichun Mo, and Leonidas~J Guibas.
\newblock Pointnet: Deep learning on point sets for 3d classification and
  segmentation.
\newblock In {\em CVPR}, 2017.

\bibitem{qi2017pointnet++}
Charles~Ruizhongtai Qi, Li~Yi, Hao Su, and Leonidas~J Guibas.
\newblock Pointnet++: Deep hierarchical feature learning on point sets in a
  metric space.
\newblock In {\em NeurIPS}, 2017.

\bibitem{chaudhuri2020learning}
Siddhartha Chaudhuri, Daniel Ritchie, Jiajun Wu, Kai Xu, and Hao Zhang.
\newblock Learning generative models of 3d structures.
\newblock In {\em Computer Graphics Forum}, 2020.

\bibitem{wald2020learning}
Johanna Wald, Helisa Dhamo, Nassir Navab, and Federico Tombari.
\newblock Learning 3d semantic scene graphs from 3d indoor reconstructions.
\newblock In {\em CVPR}, 2020.

\bibitem{liu2022bevfusion}
Zhijian Liu, Haotian Tang, Alexander Amini, Xinyu Yang, Huizi Mao, Daniela Rus,
  and Song Han.
\newblock Bevfusion: Multi-task multi-sensor fusion with unified bird's-eye
  view representation.
\newblock In {\em ICRA}, 2023.

\bibitem{liang2022bevfusion}
Tingting Liang, Hongwei Xie, Kaicheng Yu, Zhongyu Xia, Zhiwei Lin, Yongtao
  Wang, Tao Tang, Bing Wang, and Zhi Tang.
\newblock Bevfusion: A simple and robust lidar-camera fusion framework.
\newblock In {\em NeurIPS}, 2022.

\bibitem{park2021pseudo}
Dennis Park, Rares Ambrus, Vitor Guizilini, Jie Li, and Adrien Gaidon.
\newblock Is pseudo-lidar needed for monocular 3d object detection?
\newblock In {\em ICCV}, 2021.

\bibitem{guo2021liga}
Xiaoyang Guo, Shaoshuai Shi, Xiaogang Wang, and Hongsheng Li.
\newblock Liga-stereo: Learning lidar geometry aware representations for
  stereo-based 3d detector.
\newblock In {\em ICCV}, 2021.

\bibitem{chen2020dsgn}
Yilun Chen, Shu Liu, Xiaoyong Shen, and Jiaya Jia.
\newblock Dsgn: Deep stereo geometry network for 3d object detection.
\newblock In {\em CVPR}, 2020.

\bibitem{zhan2022tri}
Guanqi Zhan, Weidi Xie, Andrew Zisserman, and Coop Medianet~Innovation Center.
\newblock A tri-layer plugin to improve occluded detection.
\newblock In {\em BMVC}, 2022.

\bibitem{pan20213d}
Xuran Pan, Zhuofan Xia, Shiji Song, Li~Erran Li, and Gao Huang.
\newblock 3d object detection with pointformer.
\newblock In {\em CVPR}, 2021.

\bibitem{armeni20163d}
Iro Armeni, Ozan Sener, Amir~R Zamir, Helen Jiang, Ioannis Brilakis, Martin
  Fischer, and Silvio Savarese.
\newblock 3d semantic parsing of large-scale indoor spaces.
\newblock In {\em CVPR}, 2016.

\bibitem{qi2018frustum}
Charles~R Qi, Wei Liu, Chenxia Wu, Hao Su, and Leonidas~J Guibas.
\newblock Frustum pointnets for 3d object detection from rgb-d data.
\newblock In {\em CVPR}, 2018.

\bibitem{wilkes1992active}
D~Wilkes and JK~Tsotsos.
\newblock Active object recognition.
\newblock In {\em CVPR}, 1992.

\bibitem{palmer1999vision}
Stephen~E Palmer.
\newblock {\em Vision science: Photons to phenomenology}.
\newblock MIT press, 1999.

\bibitem{tsai2022towards}
Yao-Hung~Hubert Tsai, Hanlin Goh, Ali Farhadi, and Jian Zhang.
\newblock Towards multimodal multitask scene understanding models for indoor
  mobile agents.
\newblock In {\em ICRA}, 2023.

\bibitem{liu2022petr}
Yingfei Liu, Tiancai Wang, Xiangyu Zhang, and Jian Sun.
\newblock Petr: Position embedding transformation for multi-view 3d object
  detection.
\newblock In {\em ECCV}, 2022.

\bibitem{huang2022bevdet4d}
Junjie Huang and Guan Huang.
\newblock Bevdet4d: Exploit temporal cues in multi-camera 3d object detection.
\newblock {\em arXiv preprint arXiv:2203.17054}, 2022.

\bibitem{ma20223d}
Xinzhu Ma, Wanli Ouyang, Andrea Simonelli, and Elisa Ricci.
\newblock 3d object detection from images for autonomous driving: a survey.
\newblock {\em arXiv preprint arXiv:2202.02980}, 2022.

\bibitem{vaswani2017attention}
Ashish Vaswani, Noam Shazeer, Niki Parmar, Jakob Uszkoreit, Llion Jones,
  Aidan~N Gomez, {\L}ukasz Kaiser, and Illia Polosukhin.
\newblock Attention is all you need.
\newblock In {\em NeurIPS}, 2017.

\bibitem{wang2019dynamic}
Yue Wang, Yongbin Sun, Ziwei Liu, Sanjay~E Sarma, Michael~M Bronstein, and
  Justin~M Solomon.
\newblock Dynamic graph cnn for learning on point clouds.
\newblock {\em ACM Transactions On Graphics}, 38(5):1--12, 2019.

\bibitem{wang2021object}
Yue Wang and Justin~M Solomon.
\newblock Object dgcnn: 3d object detection using dynamic graphs.
\newblock In {\em NeurIPS}, 2021.

\bibitem{tan2019lxmert}
Hao Tan and Mohit Bansal.
\newblock Lxmert: Learning cross-modality encoder representations from
  transformers.
\newblock In {\em EMNLP}, 2019.

\bibitem{shen2019situational}
William~B Shen, Danfei Xu, Yuke Zhu, Leonidas~J Guibas, Li~Fei-Fei, and Silvio
  Savarese.
\newblock Situational fusion of visual representation for visual navigation.
\newblock In {\em ICCV}, 2019.

\bibitem{tan2022self}
Sinan Tan, Mengmeng Ge, Di~Guo, Huaping Liu, and Fuchun Sun.
\newblock Self-supervised 3d semantic representation learning for
  vision-and-language navigation.
\newblock {\em arXiv preprint arXiv:2201.10788}, 2022.

\bibitem{nilsson2021embodied}
David Nilsson, Aleksis Pirinen, Erik G{\"a}rtner, and Cristian Sminchisescu.
\newblock Embodied visual active learning for semantic segmentation.
\newblock In {\em AAAI}, 2021.

\bibitem{geiger2012we}
Andreas Geiger, Philip Lenz, and Raquel Urtasun.
\newblock Are we ready for autonomous driving? the kitti vision benchmark
  suite.
\newblock In {\em CVPR}, 2012.

\bibitem{zhudeformable}
Xizhou Zhu, Weijie Su, Lewei Lu, Bin Li, Xiaogang Wang, and Jifeng Dai.
\newblock Deformable detr: Deformable transformers for end-to-end object
  detection.
\newblock In {\em ICLR}, 2020.

\bibitem{lin2017focal}
Tsung-Yi Lin, Priya Goyal, Ross Girshick, Kaiming He, and Piotr Doll{\'a}r.
\newblock Focal loss for dense object detection.
\newblock In {\em ICCV}, 2017.

\bibitem{lin2021scene}
Xiangru Lin, Guanbin Li, and Yizhou Yu.
\newblock Scene-intuitive agent for remote embodied visual grounding.
\newblock In {\em CVPR}, 2021.

\bibitem{kingma2014adam}
Diederik~P Kingma and Jimmy Ba.
\newblock Adam: A method for stochastic optimization.
\newblock In {\em ICLR}, 2015.

\bibitem{lamb2016professor}
Alex~M Lamb, Anirudh~Goyal ALIAS PARTH~GOYAL, Ying Zhang, Saizheng Zhang,
  Aaron~C Courville, and Yoshua Bengio.
\newblock Professor forcing: A new algorithm for training recurrent networks.
\newblock In {\em NeurIPS}, 2016.

\bibitem{lin2014microsoft}
Tsung-Yi Lin, Michael Maire, Serge Belongie, James Hays, Pietro Perona, Deva
  Ramanan, Piotr Doll{\'a}r, and C~Lawrence Zitnick.
\newblock Microsoft coco: Common objects in context.
\newblock In {\em ECCV}, 2014.

\bibitem{caesar2020nuscenes}
Holger Caesar, Varun Bankiti, Alex~H Lang, Sourabh Vora, Venice~Erin Liong,
  Qiang Xu, Anush Krishnan, Yu~Pan, Giancarlo Baldan, and Oscar Beijbom.
\newblock nuscenes: A multimodal dataset for autonomous driving.
\newblock In {\em CVPR}, 2020.

\bibitem{gregory2005framework}
Arthur Gregory, Ming~C Lin, Stefan Gottschalk, and Russell Taylor.
\newblock A framework for fast and accurate collision detection for haptic
  interaction.
\newblock In {\em ACM SIGGRAPH}, 2005.

\bibitem{bergen1997efficient}
Gino van~den Bergen.
\newblock Efficient collision detection of complex deformable models using aabb
  trees.
\newblock {\em Journal of graphics tools}, 2(4):1--13, 1997.

\bibitem{huebner2008minimum}
Kai Huebner, Steffen Ruthotto, and Danica Kragic.
\newblock Minimum volume bounding box decomposition for shape approximation in
  robot grasping.
\newblock In {\em ICRA}, 2008.

\bibitem{redmon2015real}
Joseph Redmon and Anelia Angelova.
\newblock Real-time grasp detection using convolutional neural networks.
\newblock In {\em ICRA}, 2015.

\bibitem{zhou2018fully}
Xinwen Zhou, Xuguang Lan, Hanbo Zhang, Zhiqiang Tian, Yang Zhang, and Narming
  Zheng.
\newblock Fully convolutional grasp detection network with oriented anchor box.
\newblock In {\em IROS}, 2018.

\bibitem{lu2019vilbert}
Jiasen Lu, Dhruv Batra, Devi Parikh, and Stefan Lee.
\newblock Vilbert: Pretraining task-agnostic visiolinguistic representations
  for vision-and-language tasks.
\newblock In {\em NeurIPS}, 2019.

\bibitem{li2020oscar}
Xiujun Li, Xi~Yin, Chunyuan Li, Pengchuan Zhang, Xiaowei Hu, Lei Zhang, Lijuan
  Wang, Houdong Hu, Li~Dong, Furu Wei, et~al.
\newblock Oscar: Object-semantics aligned pre-training for vision-language
  tasks.
\newblock In {\em ECCV}, 2020.

\bibitem{ross2011reduction}
St{\'e}phane Ross, Geoffrey Gordon, and Drew Bagnell.
\newblock A reduction of imitation learning and structured prediction to
  no-regret online learning.
\newblock In {\em AISTATS}, 2011.

\bibitem{dosovitskiyimage}
Alexey Dosovitskiy, Lucas Beyer, Alexander Kolesnikov, Dirk Weissenborn,
  Xiaohua Zhai, Thomas Unterthiner, Mostafa Dehghani, Matthias Minderer, Georg
  Heigold, Sylvain Gelly, et~al.
\newblock An image is worth 16x16 words: Transformers for image recognition at
  scale.
\newblock In {\em ICLR}, 2020.

\bibitem{sun2020scalability}
Pei Sun, Henrik Kretzschmar, Xerxes Dotiwalla, Aurelien Chouard, Vijaysai
  Patnaik, Paul Tsui, James Guo, Yin Zhou, Yuning Chai, Benjamin Caine, et~al.
\newblock Scalability in perception for autonomous driving: Waymo open dataset.
\newblock In {\em CVPR}, 2020.

\end{thebibliography}
}

\end{document}